\documentclass[fleqn,10pt]{wlscirep}
\usepackage[symbol]{footmisc}
\usepackage{multirow}
\usepackage{graphics}
\usepackage{graphicx}
\usepackage{subcaption}
\usepackage{tabularx}
\usepackage{color}
\usepackage{amsmath}
\usepackage{hyperref}
\usepackage[ruled,vlined,linesnumbered]{algorithm2e}

\title{Characterizing Diseases from Unstructured Text:\\ A Vocabulary Driven Word2vec Approach}

\author[1, *]{Saurav Ghosh}
\author[1]{Prithwish Chakraborty}
\author[2]{Emily Cohn}
\author[2,3]{John S. Brownstein}
\author[1]{Naren Ramakrishnan}

\affil[1]{Department of Computer Science, Virginia Tech, Arlington, Vriginia, USA,}
\affil[2]{Children's Hospital Informatics Program, Boston Children's Hospital, Boston, Massachusetts, USA,}
\affil[3]{Department of Pediatrics, Harvard Medical School, Boston, Massachusetts, USA.}
\affil[*]{sauravcsvt@vt.edu}

\newcommand{\fullmodel}{\textit{Dis2Vec}}
\newcommand{\fullmodelobjective}{\textit{Dis2Vec-objective}}
\newcommand{\fullmodelsample}{\textit{Dis2Vec-sample}}
\newcommand{\fullmodelall}{\textit{Dis2Vec-combined}}
\newcommand{\skipnegative}{\textit{SGNS}}
\newcommand{\skiphierarchy}{\textit{SGHS}}
\newcommand{\bagofwords}{\textit{CBOW}}
\newcommand{\corpusdis}{\textit{$\mathcal{D}_{(d)}$}}
\newcommand{\corpusnondis}{\textit{$\mathcal{D}_{(\neg d)}$}}
\newcommand{\corpusdisnondis}{\textit{$\mathcal{D}_{(d)(\neg d)}$}}

\begin{abstract}
   Traditional disease surveillance can be augmented with a wide variety of 
real-time sources such as, news and social media. 
However, these sources
are in general unstructured and, construction of surveillance tools such as 
taxonomical correlations and trace mapping involves considerable 
human supervision.  In this paper, we motivate a disease vocabulary driven 
word2vec model ({\fullmodel}) to model diseases and constituent attributes
as word embeddings from the HealthMap news corpus.
We use these word embeddings to automatically create disease taxonomies and 
evaluate our model against corresponding human annotated taxonomies.
We compare our model accuracies against several
state-of-the art word2vec methods. Our results demonstrate that
{\fullmodel} outperforms traditional distributed vector representations in
its ability to faithfully capture taxonomical attributes across different class
of diseases such as endemic, emerging and rare.
 \end{abstract}

\begin{document}

\flushbottom
\maketitle
\thispagestyle{empty}

\section{Introduction}
  \label{sec:intro}
  Traditional disease surveillance has often relied on a multitude of reporting 
networks such as outpatient networks, on-field healthcare workers, and 
lab-based networks. Some of the most effective tools while analyzing or mapping
diseases, especially for new diseases or disease spreading to new regions, are
reliant on building disease taxonomies which can aid in early detection of outbreaks. 

In recent years, the ready availability of social and news media has led to 
services such as HealthMap~\cite{freifeld2008healthmap} which have been used to track several 
disease outbreaks from news media ranging from the flu to Ebola.
However, most of this data is unstructured and often noisy.
Annotating such corpora thus requires considerable
human oversight. 
While 
significant information about both endemic~\cite{sdm14:matrix,pg:zheng2015flu}
and rare~\cite{rekatsinasrare}
diseases can be extracted from such news corpora,
traditional text analytics methods such as lemmatization and tokenization
are often shallow and do not
retain sufficient contextual information.
More involved methods such as topic models are too computationally 
expensive for real-time worldwide surveillance
and do not provide simple semantic contexts that could be
used to comprehend the data.

\begin{figure}[ht!]
  \centering
  \includegraphics[width=0.9\linewidth]{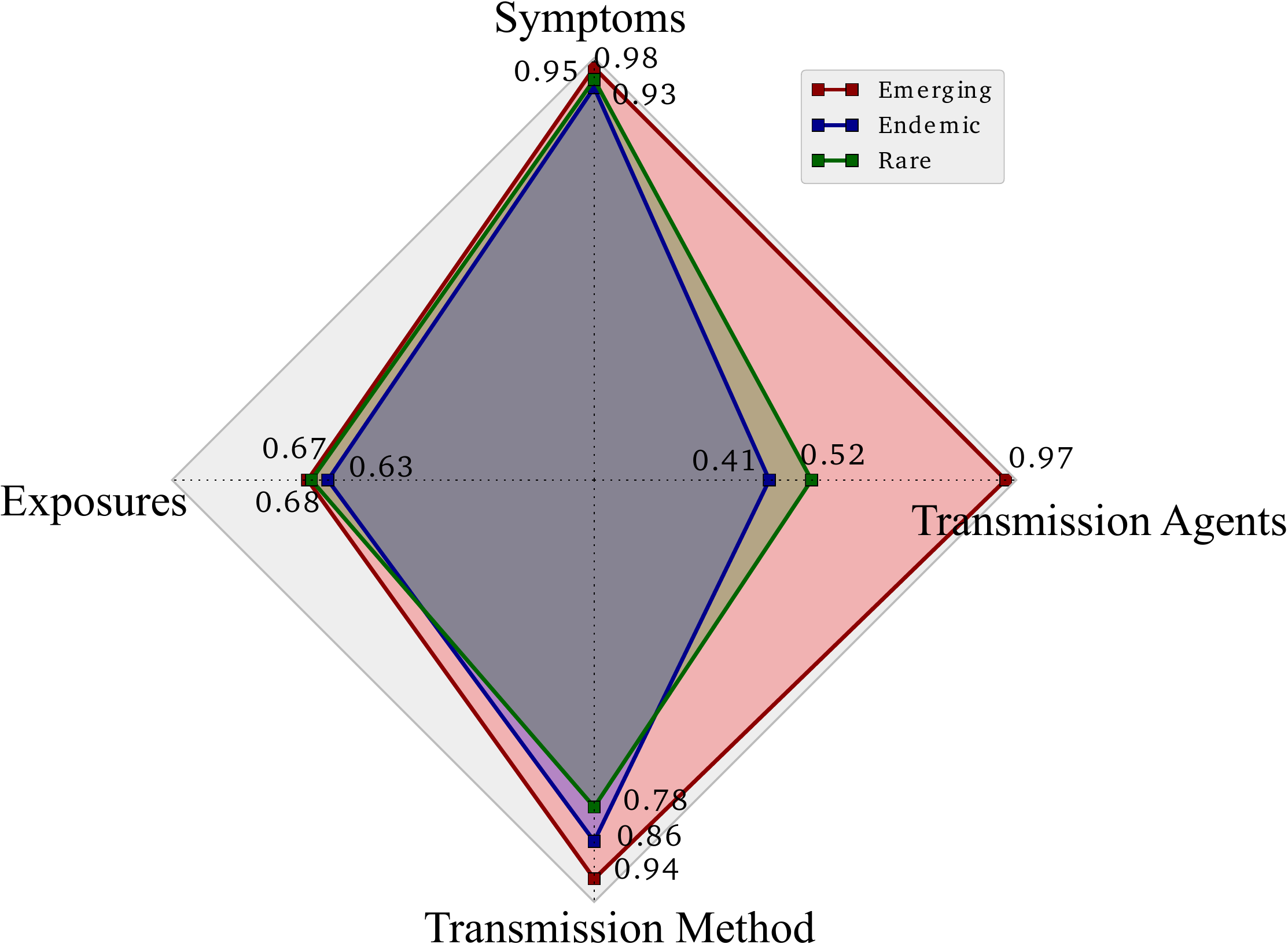}
  \caption{Comparative performance evaluation of disease specific word2vec model
  ({\fullmodel}) across the disease characterization tasks for 3 different class 
  of diseases - endemic (blue), emerging (red) and rare (green).
  The axes along the four vertices represent the modeling accuracy for the disease 
  characterization of interest viz.\ symptoms, transmission agents, transmission
  methods, and exposures.
      The area under the curve for each disease class represent the corresponding
  overall accuracy over all the characterizations. Best characterization
  performance can be seen for emerging diseases.}
  \label{fig:radar_all}
\end{figure}
In recent years, several deep learning based methods, such as word2vec 
and doc2vec, have been found to be promising in analyzing such text corpora. 
These methods once trained over a representative corpus can be readily
used to analyze new text and find semantic constructs 
(e.g. \textit{rabies}:\textit{zoonotic} = \textit{salmonella}:\textit{foodborne})
which can be useful for automated taxonomy creation.
Classical word2vec methods are generally unsupervised requiring no
domain information and as such has broad applicability. However, for highly 
specified domains (such as disease surveillance) with moderate sized corpus, 
classical methods fail to find meaningful semantic relationships.
For example, while determining the transmission method of salmonella given that
rabies is zoonotic
(i.e.\ querying \textit{rabies}:\textit{zoonotic} = \textit{salmonella}:??),
traditional word2vec methods such as skip-gram model trained on the HealthMap corpus
fail to find a meaningful answer (\textit{saintpaul}).

Motivated by this problem, in this paper we postulate a vocabulary driven 
word2vec algorithm that can find meaningful disease constructs which can be 
used towards such disease knowledge extractions. For example, for the 
aforementioned task,
vocabulary-driven word2vec algorithm generates the word \textit{foodborne},
which is more meaningful in the context of disease knowledge extraction.
Our main contributions are:\\
               
\noindent $\>\bullet$ We formulate {\fullmodel}, a vocabulary driven word2vec method 
    which is used to generate disease specific word embeddings from
    unstructured health-related news corpus.  {\fullmodel} allows domain
    knowledge in the form of pre-specified disease-related vocabulary
    $\mathcal{V}$ to supervise the discovery process of word embeddings.\\
\noindent $\>\bullet$ We use these disease specific word embeddings to generate automated
    disease taxonomies that are then evaluated against human 
    curated ones for accuracies.\\
\noindent $\>\bullet$ Finally, we evaluate the applicability of such word embeddings over
    different class of diseases - emerging, endemic and rare for different
    taxonomical characterizations.\\

\noindent \textbf{Preview of our results:} In Figure~\ref{fig:radar_all}, we
  provide a comparative performance evaluation of {\fullmodel} across the 
  disease characterization tasks for endemic, emerging and rare diseases. It
  can be seen that {\fullmodel} is best able to characterize emerging diseases.
  Specifically, it is able to capture symptoms, transmission methods and transmission
  agents, with near-perfect accuracies for emerging diseases. Such diseases (e.g.
  Ebola, H7N9) draw considerable media interest due to their unknown 
  characteristics. News articles reporting emerging outbreaks tend to focus on 
  all characteristics of such diseases - symptoms, exposures,
  transmission methods and  transmission agents. 
  However, for endemic and rare diseases, transmission agents and exposures
  are better understood, and news reports  tend to focus mainly on symptoms and 
  transmission methods. {\fullmodel} can still be applied for these class of diseases
  but with decreased accuracy for these under-represented characteristics. 
  
\section{Related Work}
  \label{sec:related}
  
The related works of interest for our problem are primarily from 
the field of neural-network based word embeddings and 
their applications in a variety of NLP tasks. 
In recent years, we have witnessed a tremendous surge of research
concerned with representing words from unstructured corpus to
dense low-dimensional vectors drawing inspirations from 
neural-network language modeling~\cite{bengio2006neural,
collobert2008unified,mnih2009scalable}.  
These representations, referred to as \textit{word embeddings}, 
have been shown to perform with considerable accuracy and ease
across a variety of linguistic tasks~\cite{baroni2014don,
collobert2011natural,turian2010word}. 

Mikolov et al.~\cite{mikolovefficient,mikolovdistributed} proposed
skip-gram model, currently a  state-of-the-art word embedding method,
which can be trained using either hierarchical softmax (SGHS)~\cite{mikolovdistributed}
or the negative sampling technique (SGNS)~\cite{mikolovdistributed}.
Skip-gram models have been found to be highly efficient in finding
word embedding templates from huge amounts of unstructured text data
and uncover various semantic and syntactic relationships.
Mikolov et al.~\cite{mikolovdistributed} also showed that the 
such word embeddings have the capability to capture linguistic regularities
and patterns. These patterns can be represented as linear translations in the
vector space. For example, vec(\textit{Madrid}) - vec(\textit{Spain}) + vec(\textit{France}) 
is closer to vec(\textit{Paris}) than any other word in their 
corpus~\cite{mikolovregular,levyregular}.

Levy et al.~\cite{levymatrixfact} analyzed the theoretical founding of 
skip-gram model and showed that the training method of SGNS can
be converted into a weighted matrix factorization and its objective induces a
implicit factorization of a shifted PMI matrix -
the well-known word-context PMI matrix~\cite{baronipmi,turneypmi}
shifted by a constant offset.
In~\cite{levyparam}, Levy et al. performed an exhaustive evaluation showing the
impact of each parameter (window size, context distribution smoothing,
sub-sampling of frequent words and others) on the performance of SGNS and other
recent word embedding methods, such as GLoVe~\cite{Glovevector}. They found
that SGNS consistently profits from larger negative samples ($ > 1$) showing
significant improvement on various NLP tasks with higher values of negative
samples.

Previous works on \textit{neural embeddings} (including the skip-gram model)
define the contexts of a word to be its linear context (words preceding and
following the target word). Levy et al.~\cite{levydependency} generalized the
skip-gram model and used syntactic contexts derived from automatically
generated dependency parse-trees. These syntactic contexts were found to
capture more functional similarities, while the bag-of-words nature of the
contexts in the original skip-gram model generates broad topical similarities.

\section{Model}
  \label{sec:model}
  \begin{figure}[t!]
  \centering
 \includegraphics[width=0.95\linewidth]{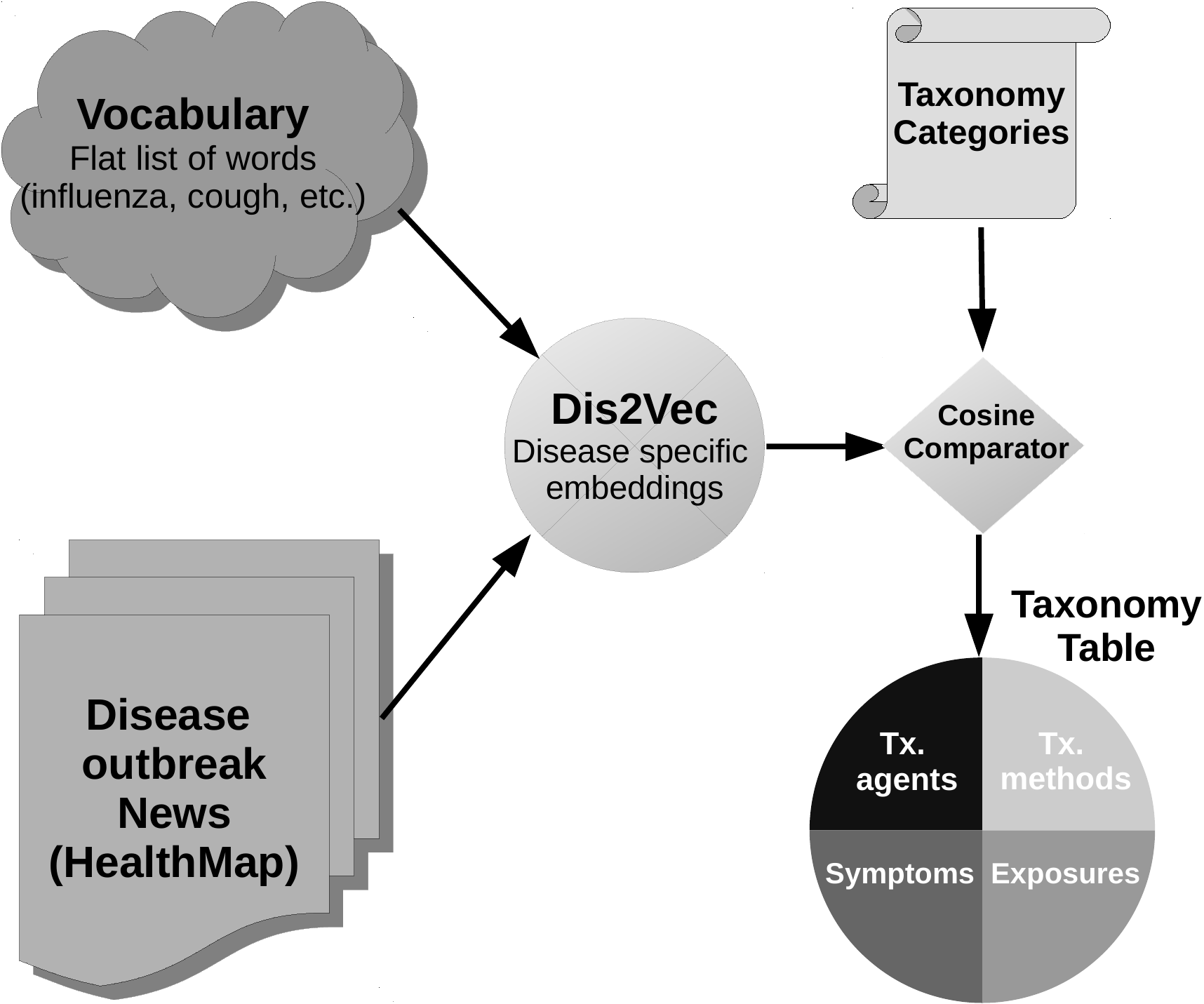}
   \caption{Automated taxonomy generation from unstructured news corpus
  (HealthMap) and a pre-specified vocabulary ($\mathcal{V}$).
  {\fullmodel} inputs these information to generate disease specific 
  word embeddings that are then passed through a cosine comparator to generate
  the taxonomy for the disease of interest.
  \label{fig:flowchart}}
\end{figure}

\subsection{Problem Overview}
Disease taxonomy generation is the process of tabulating characteristics
of diseases w.r.t.\ several pre-specified categories such as symptoms and 
transmission agents. Table~\ref{tab:taxonomy_examples} gives an example of 
taxonomy for three diseases viz.\ an emerging disease (H7N9), an endemic disease (avian influenza)
and a rare disease (plague).
Traditionally, such taxonomies are human curated - either from prior
expert knowledge or by combining a multitude of reporting sources. News reports
covering disease outbreaks can often contain disease specific information, albeit
in an unstructured way. Our aim is to generate automated taxonomy of
diseases similar to Table~\ref{tab:taxonomy_examples} using such unstructured
information from news reports. Such automated methods can greatly simplify
the process of generating taxonomies, especially for emerging diseases, and lead
to a timely dissemination of such information towards public health services.
In general, such disease related news corpus is of moderate size for deep-learning
methods and as explained in section~\ref{sec:intro}, unsupervised methods often
fail to extract meaningful information. Thus we incorporate domain knowledge in the 
form of a flat-list of disease related terms such as disease names, possible symptoms and possible
transmission methods, hereafter referred to as the 
vocabulary $\mathcal{V}$. Figure~\ref{fig:flowchart} shows the process of 
automated taxonomy generation where we employ a supervised word2vec
method referred to as {\fullmodel} which takes the following inputs - (a)
the pre-specified disease vocabulary $\mathcal{V}$ and (b) unstructured news corpus
$\mathcal{D}$ and generates embeddings for each word (including words in the
vocabulary $\mathcal{V}$) in the corpus.
Once word embeddings are generated, we employ a cosine comparator to 
create a tabular list of disease taxonomies similar to
Table~\ref{tab:taxonomy_examples}.
In this cosine comparator, to classify each
disease for a taxonomical category, we calculate the cosine similarities
between the embedding for the disease name and embeddings for all possible
words related to that category. Then, we sort these cosine similarities (in
descending order) and extract the words (higher up in the order) closer to the
disease name hereafter referred to as top words found for that category. For example,
to extract the transmission agents for \textit{plague}, we calculate
the cosine similarities between the embedding for the word \textit{plague} 
and the embeddings for all possible terms related to transmission agents 
and extract the top words by sorting the terms w.r.t.\ these similarities.
We can compare the top words found for a category with the 
human annotated words to compute the accuracy of the taxonomy
generated from word embeddings. In the next 2 subsections,
we will briefly discuss the basic word2vec model (skip-gram model with negative
sampling) followed by the detailed description of our vocabulary driven
word2vec model {\fullmodel}.

\begin{table}[!t]
 \centering
  \caption{Human curated disease taxonomy for three diseases from three 
  different class of diseases (endemic, emerging, and rare).}
  \label{tab:taxonomy_examples}
  
%\scriptsize
\begin{tabular}{|p{2.75cm}|p{2.5cm}|p{2.5cm}|p{3.0cm}|p{2.5cm}|}
\hline
Disease         & Transmission methods  & Transmission agents          & Clinical symptoms                                         & Exposures             \\ \hline
\hline
Avian influenza (endemic) & zoonotic                  & domestic animal, wild animal & Fever, cough, sore throat, diarrhea, vomiting           & animal exposure, farmer, market, slaughter        \\ \hline
H7N9 (emerging)            & zoonotic                  & domestic animal              & Fever, cough, pneumonia                                 & farmer, market, slaughter, animal exposure        \\ \hline
Plague (rare)         & vectorborne, zoonotic     & flea, wild animal            & Sore, fever, headache, muscle ache, vomiting, nausea    & animal exposure, veterinarian, farmer       \\ \hline
\end{tabular}
 \end{table}

\subsection{Basic Word2vec Model}
\label{sec:SGNS_model}
 In this section, we present a brief description of {\skipnegative} - the skip-gram model introduced in~\cite{mikolovefficient}
 trained using the negative sampling procedure in~\cite{mikolovdistributed}. The objective of the skip-gram model is to 
  infer word embeddings that will be relevant for predicting the surrounding words in a sentence or a document.
  The skip-gram model can also be trained using Hierarchical Softmax method as shown in~\cite{mikolovdistributed}.
  
\subsubsection{Setting and Notation} 
The inputs to the skip-gram model are a corpus of words $w \in \mathcal{W}$ and their corresponding
contexts $c \in \mathcal{C}$ where $\mathcal{W}$ and $\mathcal{C}$ are the word and context vocabularies. In {\skipnegative}, the contexts of word $w_{i}$ are defined
  by the words surrounding it in an $L$-sized context window $w_{i-L},\ldots,w_{i-1},w_{i+1},\ldots,w_{i+L}$. Therefore, the corpus can be 
  transformed into a collection of observed context and word pairs as $\mathcal{D}$. The notation $\#(w,c)$ represents the number of 
  times the pair $(w,c)$ occurs in $\mathcal{D}$. Therefore, $\#(w) = \sum_{c \in \mathcal{C}} (w,c)$ and $\#(c) = \sum_{w \in \mathcal{W}} (w,c)$ where $\#(w)$ and $\#(c)$ are 
  the total number of times $w$ and $c$ occurred in $\mathcal{D}$. Each word $w \in \mathcal{W}$ corresponds to a vector $\mathbf{w} \in R^{T}$
  and similarly, each context $c \in \mathcal{C}$ is represented as a vector $\mathbf{c} \in R^{T}$, where $T$ is the dimensionality of the
  word or context embedding. The entries in the vectors are the latent parameters to be learned. 

  \subsubsection{Objective of {\skipnegative}}
  {\skipnegative} tries to maximize the probability whether a single word-context 
  pair $(w,c)$ was generated from the observed corpus $\mathcal{D}$. 
  Let $P(\mathcal{D} = 1|w, c)$ refers to the probability that $(w, c)$ was generated from the corpus, 
  and $P(\mathcal{D} = 0|w, c) = 1 - P(\mathcal{D} = 1|w, c)$ the probability that $(w, c)$ was not. The objective 
  function for a single $(w,c)$ pair is modeled as:

  \begin{equation}
    \label{eq:SGNSsingleobj}
    \begin{array}{l}
      P(\mathcal{D}=1|w,c) = \sigma(\mathbf{w} \cdot \mathbf{c}) = \frac{1}{1+e^{-\mathbf{w}\cdot\mathbf{c}}}
    \end{array}
  \end{equation}

  \noindent where $\mathbf{w}$ and $\mathbf{c}$ are the $T$-dimensional latent parameters or vectors 
  to be learned.
 
  The objective of the negative sampling is to maximize $P(\mathcal{D} = 1|w, c)$ 
  for observed $(w, c)$ pairs while minimizing $P(\mathcal{D} = 0|w, c)$ for randomly 
  sampled "negative" contexts (hence the name "negative
  sampling"), under the assumption that randomly selecting a context for a given word will tend 
  to generate an unobserved $(w, c)$ pair. {\skipnegative}'s objective for a single $(w, c)$ observation is then:

  \begin{equation}
    \label{eq:SGNSsingleobj}
    \begin{array}{l}
      \log\sigma(\mathbf{w} \cdot \mathbf{c}) + k\cdot E_{c_{N} \sim P_{\mathcal{D}}}[\log\sigma(-\mathbf{w}\cdot\mathbf{c_{N}})]
    \end{array}
  \end{equation}

  \noindent where $k$ is the number of "negative" samples and $c_{N}$ is the sampled 
  context, drawn according to the smoothed unigram distribution $P_{\mathcal{D}}(c) = \frac{\#(c)^{\alpha}}{\sum_{c}\#(c)^{\alpha}}$ 
  where $\alpha=0.75$ is the smoothing parameter.

  The objective of $SGNS$ is trained in an online fashion using stochastic gradient updates over the observed
  pairs in the corpus $\mathcal{D}$. The global objective then sums over the observed $(w, c)$ pairs in the corpus:

  \begin{equation}
    \label{eq:SGNSfullobj}
    \begin{array}{l}
      l_{\skipnegative} = \sum_{(w,c) \in \mathcal{D}}\biggl(\log\sigma(\mathbf{w} \cdot \mathbf{c}) + k\cdot E_{c_{N} \sim P_{\mathcal{D}}}[\log\sigma(-\mathbf{w}\cdot\mathbf{c_{N}})]\biggr)
    \end{array}
  \end{equation}

  \noindent Optimizing this objective will have a tendency to generate similar embeddings 
  for observed word-context pairs, while scattering unobserved pairs in the vector space. 
  Intuitively, words that appear in similar contexts or tend to appear in the contexts of 
  each other should have similar embeddings.

  \subsection{Disease Specific Word2vec Model ({\fullmodel})}
  \label{sec:our_model}
  In this section, we introduce {\fullmodel}, a disease specific word2vec model 
  whose objective is to generate word embeddings which will be useful for automatic disease taxonomy creation 
  given an input unstructured corpus $\mathcal{D}$. We used a pre-specified disease-related vocabulary $\mathcal{V}$ 
  (domain information) to guide the discovery process of word embeddings in {\fullmodel}. The input corpus $\mathcal{D}$ 
  consists of a collection of $(w, c)$ pairs. Based on $\mathcal{V}$, we can categorize the $(w,c)$ pairs into three types as shown below:\\

      \noindent $\>\bullet$  {\corpusdis} $= \{(w,c) \colon w \in \mathcal{V} \wedge c \in \mathcal{V} \}$, i.e.\ both the word $w$ and the context $c$ are in $\mathcal{V}$ \\
   \noindent $\>\bullet$  {\corpusnondis} $= \{(w,c) \colon w \notin \mathcal{V} \wedge c \notin \mathcal{V} \}$, i.e.\ neither the word $w$ nor the context $c$ are in $\mathcal{V}$ \\
   \noindent $\>\bullet$  {\corpusdisnondis} $= \{(w,c) \colon w \in \mathcal{V} \oplus  c \in \mathcal{V} \}$, i.e.\ either the word $w$ is in $\mathcal{V}$ or the context $c$ is in $\mathcal{V}$ but both cannot be in $\mathcal{V}$ \\
   
   \noindent Therefore, the input corpus $\mathcal{D}$ can be represented as $\mathcal{D} =$ {\corpusdis} + {\corpusnondis} + {\corpusdisnondis}. Each of these categories of
   $(w,c)$ pairs needs special consideration while generating disease specific embeddings.

   \subsubsection{Vocabulary Driven Negative Sampling}
     \label{sec:first_category}

     The first category ({\corpusdis}) of $(w,c)$ pairs, where both $w$ and $c$ are in $\mathcal{V}$ 
     ($w \in \mathcal{V} \wedge c \in \mathcal{V}$), is of prime importance in generating disease specific word embeddings.
     Our first step in generating such embeddings is to maximize $\log\sigma(\mathbf{w} \cdot \mathbf{c})$ in order to achieve 
     similar embeddings for these disease word-context pairs. Apart from maximizing the dot products, following
     classical approaches~\cite{mikolovdistributed}, negative sampling is also required to generate robust embeddings.
     In {\fullmodel}, we adopt a vocabulary ($\mathcal{V}$) driven negative sampling for these disease word-context pairs.
     In this vocabulary driven approach, instead of random sampling we sample negative 
     examples ($c_{N}$) from the set of non-disease contexts, i.e.\ contexts which are not in $\mathcal{V}$ ($c \notin \mathcal{V}$). 
     This targeted sampling of negative contexts will ensure dissimilar embeddings of disease words ($w \in \mathcal{V}$) and non-disease 
     contexts ($c \notin \mathcal{V}$), thus scattering them in the vector space. However, sampling negative examples only 
     from the set of non-disease contexts may lead to overfitting and thus, we introduce a sampling 
     parameter $\pi_{s}$ which controls the  probability of drawing a \textit{negative} example from non-disease 
     contexts ($c \in \mathcal{V}$) versus disease contexts ($c \in \mathcal{V}$). {\fullmodel}'s objective for $(w,c) \in {\corpusdis}$ is shown below in
     equation~\ref{eq:VSGNSfirstobj}.
     \begin{align}
    \label{eq:VSGNSfirstobj}
      l_{\corpusdis} &= \sum_{(w,c) \in {\corpusdis}} \biggl( \log\sigma(\mathbf{w} \cdot \mathbf{c}) \\ 
      &+ k\cdot [P(x_{k} < \pi_{s})E_{c_{N} \sim P_{D_{c \notin \mathcal{V}}}}[\log\sigma(-\mathbf{w}\cdot\mathbf{c_{N}})]  \nonumber\\
      &+ P(x_{k} \geq \pi_{s})E_{c_{N} \sim P_{D_{c \in \mathcal{V}}}}[\log\sigma(-\mathbf{w}\cdot\mathbf{c_{N}})]] \biggr) \nonumber
    \end{align}
    
    \noindent where $x_{k} \sim U(0,1)$, U(0,1) being the uniform distribution on the interval [0,1]. If $x_{k} < \pi_{s}$, 
    we sample a negative context $c_{N}$ from the unigram distribution $P_{D_{c \notin \mathcal{V}}}$ where $D_{c \notin \mathcal{V}}$ is the collection
    of $(w,c)$ pairs for which $c \notin \mathcal{V}$ and $P_{D_{c \notin \mathcal{V}}} = \frac{\#(c)^{\alpha}}{\sum_{c \notin \mathcal{V}}\#(c)^{\alpha}}$ 
    where $\alpha$ is the smoothing parameter. For values of $x_{k} \geq \pi_{s}$, we sample $c_{N}$ from the unigram distribution 
    $P_{D_{c \in \mathcal{V}}}$ and $P_{D_{c \in \mathcal{V}}} = \frac{\#(c)^{\alpha}}{\sum_{c \in V}\#(c)^{\alpha}}$. 
    Therefore, optimizing the objective in equation~\ref{eq:VSGNSfirstobj} will have a tendency to generate disease specific word embeddings for values 
    of $\pi_{s} \geq 0.5$ due to the reason that higher number of negative contexts ($c_{N}$) will be sampled from the set of non-disease contexts 
    ($c \notin \mathcal{V}$) with $\pi_{s} \geq 0.5$.

    \subsubsection{Out-of-vocabulary Objective Regularization}
     \label{sec:second_category}
     The second category ({\corpusnondis}) of $(w,c)$ pairs consists of those pairs for which both $w$ and $c$ are not in $\mathcal{V}$ ($w \notin \mathcal{V}
     \wedge c \notin \mathcal{V}$). These pairs are uninformative to us in generating disease specific word embeddings since 
     both $w$ and $c$ are not a part of $\mathcal{V}$. However, minimizing their dot products will scatter these pairs 
     in the embedding space and thus, a word $w \notin \mathcal{V}$ can have similar embeddings (or, get closer) to a word 
     $w \in \mathcal{V}$ which should be avoidable in our scenario. Therefore, we need to maximize $\log\sigma(\mathbf{w} \cdot \mathbf{c})$ 
     for these $(w,c)$ pairs in order to achieve similar (or, closer) embeddings. We adopt the basic objective function of {\skipnegative}  
     for $(w,c) \in$ {\corpusnondis} as shown below in equation~\ref{eq:VSGNSsecondobj}. 

    \begin{equation}
    \label{eq:VSGNSsecondobj}
          l_{\corpusnondis} = \sum\limits_{(w,c) \in {\corpusnondis}}\biggl(\log\sigma(\mathbf{w} \cdot \mathbf{c}) 
                        + k\cdot E_{c_{N} \sim P_{D}}[\log\sigma(-\mathbf{w}\cdot\mathbf{c_{N}})] \biggr) 
         \end{equation}

     \subsubsection{Vocabulary Driven Objective Minimization}
        \label{sec:third_category}
        Lastly, the third category ({\corpusdisnondis}) consists of $(w,c)$ pairs where either $w$ is in $\mathcal{V}$ or $c$ is in $\mathcal{V}$ 
        ($w \in \mathcal{V} \oplus c \in \mathcal{V}$) 
        but both cannot be in $\mathcal{V}$. Consider an arbitrary $(w,c)$ pair belonging to {\corpusdisnondis}. As per the objective (equation~\ref{eq:SGNSfullobj}) 
     of {\skipnegative}, two words are similar to each other if they share the same contexts or if they tend to appear in the contexts
     of each other (and preferably both). If $w \in \mathcal{V}$ and $c \notin \mathcal{V}$, then maximizing  $\log\sigma(\mathbf{w} \cdot \mathbf{c})$ 
     will have the tendency to generate similar embeddings for the disease word $w \in \mathcal{V}$ and non-disease words $\notin \mathcal{V}$ 
     which share the same non-disease context $c \notin \mathcal{V}$. On the other word, if $c \in \mathcal{V}$ 
     and $w \notin \mathcal{V}$, then maximizing  $\log\sigma(\mathbf{w} \cdot \mathbf{c})$ will drive the embedding of the non-disease word $w \notin \mathcal{V}$ 
     closer to the embeddings of disease words $\in \mathcal{V}$ sharing the same disease context $c \in \mathcal{V}$. Therefore, we posit 
     that the dot products for this category of $(w,c)$ pairs should be minimized, i.e.\ the objective $\log\sigma(-\mathbf{w} \cdot \mathbf{c})$ should be optimized 
     in order to ensure dissimilar embeddings for these $(w,c)$ pairs. However, minimizing the dot products of all such word-context pairs may lead to over-penalization and thus we introduce an 
     objective selection parameter $\pi_{o}$ which controls the probability of selecting  $\log\sigma(-\mathbf{w} \cdot \mathbf{c})$ versus $\log\sigma(\mathbf{w} \cdot \mathbf{c})$. 
     The objective for $(w,c) \in$ {\corpusdisnondis} is shown below in equation~\ref{eq:VSGNSthirdobj}.

     \begin{align}
    \label{eq:VSGNSthirdobj}
          l_{\corpusdisnondis} &= \sum_{(w,c) \in {\corpusdisnondis}}\biggl(P(z < \pi_{o})\log\sigma(-\mathbf{w}\cdot\mathbf{c}) \\
                           &+ P(z \geq \pi_{o})\log\sigma(\mathbf{w}\cdot\mathbf{c})\biggr) \nonumber
        \end{align}
 
    \noindent where $z \sim U(0,1)$, U(0,1) being the uniform distribution over the interval [0,1].
    If $z < \pi_{o}$, $\log\sigma(-\mathbf{w} \cdot \mathbf{c})$ gets optimized, otherwise 
    {\fullmodel} optimizes $\log\sigma(\mathbf{w} \cdot \mathbf{c})$. Therefore, optimizing the objective in equation~\ref{eq:VSGNSthirdobj}
    will have a tendency to generate disease specific embeddings with values of $\pi_{o} \geq 0.5$ due to the reason that  
    the objective $\log\sigma(-\mathbf{w} \cdot \mathbf{c})$ will be selected for optimization with 
    a higher probability over $\log\sigma(\mathbf{w} \cdot \mathbf{c})$.
    \\
    Finally, the overall objective of {\fullmodel} comprising all three categories of $(w,c)$ pairs can 
    be defined as below.

    \begin{equation}
    \label{eq:VSGNSallobj}
    \begin{array}{l}
      l_{\fullmodel} = l_{\corpusdis} + l_{\corpusnondis} + l_{\corpusdisnondis}    
    \end{array}
    \end{equation}

    \noindent Similar to {\skipnegative}, the objective in equation~\ref{eq:VSGNSallobj} is trained in an online fashion
    using stochastic gradient updates over the three categories of $(w,c)$ pairs.

\begin{algorithm}
    %\scriptsize
    \DontPrintSemicolon
    \SetKwInOut{Input}{Input}\SetKwInOut{Output}{Output}
    \Input{Unstructured corpus $\mathcal{D} = \{(w,c)\}$, $\mathcal{V}$}
    \Output{word embeddings $\mathbf{w} \forall w \in \mathcal{W}$, column embeddings $\mathbf{c} \forall c \in \mathcal{C}$}
    Categorize $\mathcal{D}$ into 3 types: $\corpusdis~= \{(w,c):  w \in \mathcal{V} \wedge c \in \mathcal{V}\}$, $\corpusnondis~= \{(w,c) : w \notin \mathcal{V} \wedge c \notin \mathcal{V}\}$,  
    $\corpusdisnondis~= \{(w,c) : w \in \mathcal{V} \oplus c \in \mathcal{V}\} $\;
    \For{each $(w,c) \in \mathcal{D}$}{
    \If{$(w,c) \in {\corpusdis}$}{
            train the $(w,c)$ pair using the objective in equation~\ref{eq:VSGNSfirstobj}\;
            }
            \ElseIf{$(w,c) \in {\corpusnondis}$}{
            train the $(w,c)$ pair using the objective in equation~\ref{eq:VSGNSsecondobj}\;
            }
      \Else{
            train the $(w,c)$ pair using the objective in equation~\ref{eq:VSGNSthirdobj}\;
            }
          }
    \caption{{\fullmodel} model\label{al:Dis2Vec}}
\end{algorithm}

    \subsection{Parameters in {\fullmodel}}

    {\fullmodel} inherits all the parameters of {\skipnegative}, such as dimensionality ($T$) of the word embeddings, 
    window size ($L$), number of negative samples ($k$) and context distribution smoothing ($\alpha$). It also
    introduces two new parameters - the objective selection parameter ($\pi_{o}$) and the sampling parameter ($\pi_{s}$).
    The explored values for each of the aforementioned parameters are shown in Table~\ref{tab:beneficial_param}.

\section{Experimental Evaluation}
  \label{sec:expts}
  We evaluated {\fullmodel} against several state-of-the art methods. 
In this section, we first provide a brief description of our experimental
setup, including the disease news corpus, human annotated taxonomy and the
domain information used as the vocabulary $\mathcal{V}$ for the process.
We present our experimental findings in 
Section~\ref{sec:results} where we have compared our model against several
baselines and also explore its applicability to emerging diseases.

\subsection{Experimental Setup}
\label{sec:expt}

\subsubsection{Corpus}

We collected a dataset corresponding to a corpus of public health-related
news articles in English extracted from HealthMap~\cite{freifeld2008healthmap}, a prominent
online aggregator of news articles from all over the world for disease outbreak
monitoring and real-time surveillance of emerging public health threats. Each article  
contains the following information - textual content, disease tag, reported date and location 
information in the form of (lat, long) coordinates. The articles were reported during the time period 
2010 to 2014 and correspond to locations from all over the world. The textual content of each article 
was pre-processed by sentence splitting, tokenization and lemmatization via BASIS
Technologies' Rosette Language Processing (RLP) tools~\cite{narenembers}. After
pre-processing, the corpus consisting of 124850 articles was found to contain
1607921 sentences, spanning 52679298 words. Words that appeared less than 5
times in the corpus were ignored, resulting in a vocabulary of 91178 words.
  
\subsubsection{Human Annotated Taxonomy} 
Literature reviews were conducted for each of the 39 infectious diseases of
interest in order to make classifications for transmission methods,
transmission agents, clinical symptoms and exposures or risk factors. 

Methods of transmission were first classified into 8 subcategories -
\textit{direct contact}, \textit{droplet}, \textit{airborne},
\textit{zoonotic}, \textit{vectorborne}, \textit{waterborne},
\textit{foodborne}, and \textit{environmental}. For many diseases, multiple
subcategories of transmission methods could be assigned.  Transmission
agents were classified into 8 subcategories - \textit{wild animal},
\textit{fomite}, \textit{fly}, \textit{mosquito}, \textit{bushmeat},
\textit{flea}, \textit{tick} and \textit{domestic animal}.  The category of
clinical symptoms was broken down into 8 subcategories: \textit{general},
\textit{gastrointestinal}, \textit{respiratory}, \textit{nervous system},
\textit{cutaneous}, \textit{circulatory}, \textit{musculoskeletal}, and
\textit{urogenital}. A full list of the symptoms within each subcategory can be
found in Table~\ref{tab:taxonomy_symptoms}.  For disease exposures or risk
factors, 7 subcategories were assigned based on those listed/most commonly
reported in the literature. The subcategories include: \textit{healthcare
facility} , \textit{healthcare worker}, \textit{schoolchild}, \textit{mass
gathering}, \textit{travel}, \textit{animal exposure}, and \textit{weakened
immune system}. The \textit{animal exposure} category was further broken down
into \textit{farmer}, \textit{veterinarian}, \textit{market} and
\textit{slaughter}. For some diseases, there were no risk factors listed, and
for other diseases, multiple exposures were assigned. 
 
\begin{table}[!ht]
  \centering
  \small
  \caption{Symptom categories and corresponding words.}
  \label{tab:taxonomy_symptoms}
  
%\scriptsize
\begin{tabularx}{0.49\textwidth}{|l|X|}
\hline
Symptom Category         & Words                                                                            \\ 
\hline
\hline
General          & Fever, chill, weight loss, fatigue, lethargy, headache                              \\ \hline
Gastrointestinal & Abdominal pain, nausea, diarrhea, vomiting                                          \\ \hline
Respiratory      & Cough, runny nose, sneezing, chest pain, sore throat, pneumonia, dyspnea                      \\ \hline
Nervous system   & Mental status, paralysis, paresthesia, encephalitis, meningitis                                  \\ \hline
Cutaneous        & Rash, sore, pink eye                                                                 \\ \hline
Circulatory      & Hemorrhagic                                                                            \\ \hline
Musculoskeletal  & Joint pain, muscle pain, muscle ache                                                       \\ \hline
\end{tabularx}
 \end{table}

\begin{figure}[h!]
  \centering
  \includegraphics[width=0.9\linewidth]{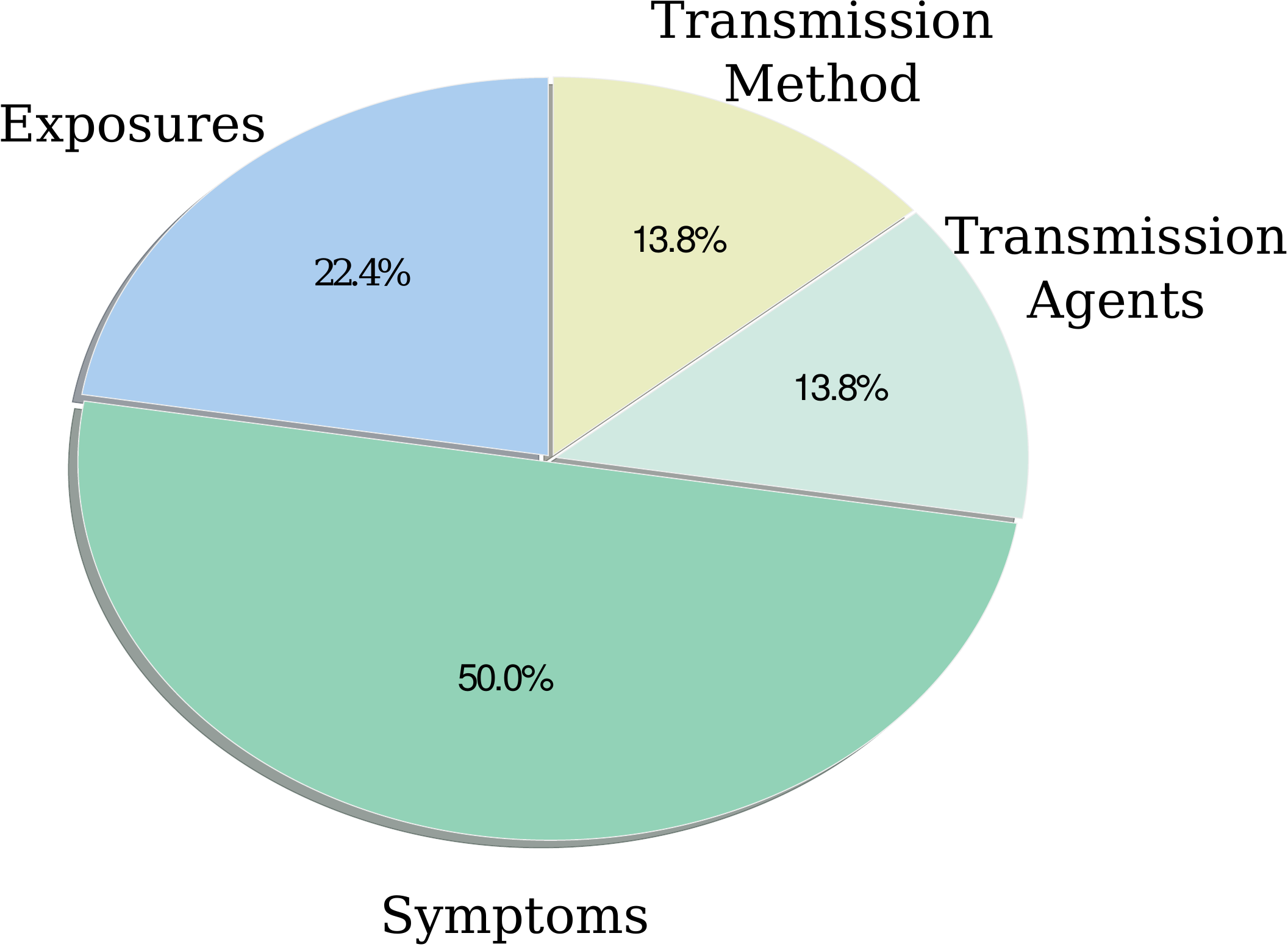}
    \caption{Distribution of word counts corresponding to each taxonomical category in the disease vocabulary ($\mathcal{V}$).
  Words related to clinical symptoms constitute the majority of $\mathcal{V}$ with relatively much smaller
  percentages of terms related to exposures, transmission agents and transmission methods}
  \label{fig:vocab_dist}
\end{figure}

\subsubsection{Disease Vocabulary $\mathcal{V}$}

Disease vocabulary $\mathcal{V}$ is provided as prior knowledge to {\fullmodel} in order to generate
disease specific word embeddings as explained in section~\ref{sec:our_model}. $\mathcal{V}$ is represented 
by a flat list of disease-related terms consisting of disease names (\textit{influenza}, \textit{h7n9}, \textit{plague}, \textit{ebola}, etc.), 
all possible words related to transmission methods(\textit{vectorborne}, \textit{foodborne}, \textit{waterborne}, etc.), 
all possible words related to transmission agents (\textit{flea}, \textit{domestic animal}, \textit{mosquito}, etc.), all possible words
related to clinical symptoms (\textit{fever}, \textit{nausea}, \textit{paralysis}, \textit{cough}, \textit{headache}, etc.) and all possible words related to 
exposures or risk factors (\textit{healthcare facility}, \textit{slaughter}, \textit{farmer}, etc.). 
Total number of words in $\mathcal{V}$ is found to be 103. In Figure~\ref{fig:vocab_dist}, 
we show the distribution of word counts associated with different taxonomical categories in the disease vocabulary ($\mathcal{V}$).
As depicted in Figure~\ref{fig:vocab_dist}, half of the words in $\mathcal{V}$ are terms related to clinical symptoms followed
by exposures or risk factors ($22.4 \%$), transmission methods ($13.8 \%$) and transmission agent(s) ($13.8 \%$).

\subsubsection{Baselines}
\label{sec:models}
We compared the following baseline models with {\fullmodel} on the four disease
characterization tasks.\\

\noindent $\bullet$ \textbf{\skipnegative}: Unsupervised skip-gram model with
  negative sampling~\cite{mikolovdistributed} described in section~\ref{sec:SGNS_model}.\\
\noindent $\bullet$ \textbf{\skiphierarchy}: skip-gram model trained using
  the hierarchical softmax algorithm~\cite{mikolovdistributed} instead of negative sampling. 
  \\
\noindent $\bullet$ \textbf{\bagofwords}: Continuous bag-of-words model described
  in~\cite{mikolovefficient}. Unlike skip-gram models, the training objective
  of the {\bagofwords} model is to correctly predict the target word given its
  contexts (surrounding words). $CBOW$ is denoted as a bag-of-words model as
  the order of words in the contexts does not have any impact on the model.\\

All models (both baselines and {\fullmodel}) were trained on the HealthMap corpus 
using a $T$-dimensional word embedding via gensim's word2vec software~\cite{rehurek_lrec}. 
We explored a large space of parameters for each model.  In Table~\ref{tab:beneficial_param}, we provide the
list of parameters, the explored values for each parameter and the applicable
models corresponding to each parameter. Apart from the parameters listed in 
Table~\ref{tab:beneficial_param}, we also applied the sub-sampling technique 
developed by Mikolov et al.~\cite{mikolovdistributed} to each model in order to counter the imbalance between
common words (such as, \textit{is}, \textit{of}, \textit{the}, \textit{a}, etc.) and rare words.
In the context of NLP, these common words are referred to as \textit{stop words}.  For more details on the sub-sampling techniques, 
please see Mikolov et al.~\cite{mikolovdistributed}. 
Our initial experiments (not reported) demonstrated that both the baselines and {\fullmodel} showed improved results on the 
disease characterization tasks with sub-sampling versus without sub-sampling.

\subsubsection{Accuracy Metric}

We evaluate the automatic taxonomy generation methods such that for 
a taxonomical characteristic of a disease, models that generate
similar set of terms (top words) as the human annotated ones are more preferable.
As such, we use cosine similarity in a min-max setting 
between the aforementioned sets for a particular characterization
category as our accuracy metric. The overall accuracy of a model for a category 
can be found by averaging the accuracy values across all diseases of interest.
This is a bounded metric (between $0$ and $1$)
where higher values indicate better model performance. 
We can formalize the metric as follows. Let $D$ be the disease and 
$C$ be the taxonomical category under investigation.
Furthermore, let $C_{1}, C_{2},\cdots,C_{N}$ be all
possible terms or words related to $C$ and $H_{1}, H_{2},\cdots,H_{M}$ be 
the human annotated words. Then the characterization accuracy corresponding 
to category $C$ and disease $D$ is given below in equation~\ref{eq:accmetric}.
\begin{equation}
  \scriptsize
  Accuracy(C, D) = \frac{1}{M}\sum_{j=1}^{M}\frac{cosine(\mathbf{D}, \mathbf{H_{j}})-\min_{i}cosine(\mathbf{D}, \mathbf{C_{i}})}{\max_{i}cosine(\mathbf{D}, \mathbf{C_{i}})-\min_{i}cosine(\mathbf{D}, \mathbf{C_{i}})}
  \label{eq:accmetric}
\end{equation}

\noindent where $\mathbf{D}$, $\mathbf{H_{j}}$ and $\mathbf{C_{i}}$ represent
the word embeddings for $D$, $H_{j}$ and $C_{i}$.  $\min_{i}cosine(\mathbf{D},
\mathbf{C_{i}})$ and $\max_{i}cosine(\mathbf{D}, \mathbf{C_{i}})$ represent the
maximum and minimum cosine similarity values between $\mathbf{D}$ and the word
embeddings of the terms related to $C$. Therefore, equation~\ref{eq:accmetric}
indicates that if the human annotated word $\mathbf{H_{j}}$ is among the top
words found by the word2vec model for the category $C$, then the ratio in the
numerator is high leading to high accuracy and vice versa.

\subsection{Results}
  \label{sec:results}
  
In this section we try to ascertain the efficacy and the applicability of 
{\fullmodel} by investigating some of the pertinent questions related to 
the problem of disease characterization.
\begin{enumerate}
  \item \textbf{Sample-vs-objective: which is the better method to incorporate
    disease vocabulary information into {\fullmodel}?}
  \item \textbf{Does disease vocabulary information improve disease
     characterization? }
  \item \textbf{What are beneficial parameter configurations for characterizing 
    diseases?}
  \item \textbf{Importance of taxonomical categories - how should we construct
    the disease vocabulary? }
  \item \textbf{Can {\fullmodel} be applied to characterize emerging, endemic 
    and rare diseases?}
\end{enumerate}

\par{\textbf{Sample-vs-objective: which is the better method to incorporate
    disease vocabulary information into {\fullmodel}?}
As described in Section~\ref{sec:model}, there are primarily two
different ways by which disease vocabulary information ($\mathcal{V}$) guides the generation
of embeddings for {\fullmodel}~(a) by modulating negative sampling parameter ($\pi_{s}$) 
for disease word-context pairs ($(w,c) \in {\corpusdis}$) referred to as {\fullmodelsample} 
and (b) by modulating the objective selection parameter ($\pi_{o}$) 
for non-disease words or non-disease contexts ($(w,c) \in {\corpusdisnondis}$) referred to as 
{\fullmodelobjective}. We investigate the importance of these two strategies by 
comparing the accuracies for each strategy individually  
({\fullmodelsample}~and {\fullmodelobjective}) as well as combined together
({\fullmodelall}) under the best parameter configuration for a particular task 
in Table~\ref{tab:model_accuracy_best_parameter}. As can be seen, no single strategy is best across 
all tasks. Henceforth, we select the best performing strategy for a particular task as our {\fullmodel} 
in the next Table~\ref{tab:model_accuracy_baseline_best_parameter}.}

\begin{table}[!ht]
  \centering 
  \caption{Comparative performance evaluation of {\fullmodelall} against
  {\fullmodelobjective} and {\fullmodelsample} across the 4 characterization tasks
  under the best parameter configuration for that model and task combination. The 
  value in each cell represents the overall accuracy across the 39 diseases 
  for that particular model and characterization task. We use equation~\ref{eq:accmetric} 
  as the accuracy metric in this table.}
  \label{tab:model_accuracy_best_parameter}
  %\scriptsize
\begin{tabular}{|p{5.0cm}|p{3.1cm}|p{3.1cm}|p{3.1cm}|}
  \hline
  Characterization tasks       & {\fullmodelsample}   & {\fullmodelobjective} & {\fullmodelall}  \\ 
  \hline
  \hline
  Symptoms                 & 0.635      & {\bf 0.945} & 0.940  \\ \hline
  Exposures                & 0.590 & 0.540 & {\bf 0.597} \\ \hline
  Transmission methods   & {\bf 0.794} & 0.754 & 0.734 \\ \hline
  Transmission agents      & 0.505       & 0.506 & {\bf 0.516}    \\ \hline
  Overall average accuracy & 0.631 & 0.686 & {\bf 0.697} \\ \hline
\end{tabular}
 
 \end{table}

\begin{table}[!ht]
  \centering 
  \caption{Comparative performance evaluation of {\fullmodel} against
  {\skipnegative}, {\skiphierarchy} and {\bagofwords} across the 4 characterization
  tasks under the best parameter configuration for that model and task
  combination. The value in each cell represents the overall accuracy across the 39 diseases 
  for that particular model and characterization task. We use equation~\ref{eq:accmetric} 
  as the accuracy metric in this table.}
  \label{tab:model_accuracy_baseline_best_parameter}
  %\scriptsize
\begin{tabular}{|m{4.0cm}|m{1.8cm}|m{1.8cm}|m{1.8cm}|m{1.9cm}|}
  \hline
  Characterization tasks       & {\bagofwords} & {\skiphierarchy}   & {\skipnegative}   & {\fullmodel}  \\ 
  \hline
  \hline
  Symptoms            & 0.498 & 0.560 & 0.620 & {\bf 0.945}   \\ \hline
  Exposures           & 0.383 & 0.498 & {\bf 0.605} & 0.597 \\ \hline
  Transmission methods & 0.481 & 0.765 & 0.792 & {\bf 0.794} \\ \hline
  Transmission agents & 0.274 & 0.466 & 0.498 & {\bf 0.516} \\ \hline
  Overall average accuracy & 0.409 & 0.572 & 0.629 & {\bf 0.713} \\ \hline
\end{tabular}

 \end{table}

\par{\textbf{Does disease vocabulary information improve disease characterization?} 
{\fullmodel} was designed to incorporate disease vocabulary information in order to guide the generation
of disease specific word embeddings. To evaluate the importance of such
vocabulary information in {\fullmodel}, we compare the performance of {\fullmodel}
against the baseline word2vec models described in 
section~\ref{sec:models} under the best parameter configuration for a particular task. These baseline models 
do not permit incorporation of any vocabulary information due to their unsupervised nature. Table~\ref{tab:model_accuracy_baseline_best_parameter} 
presents the accuracy of the models for the 4 disease characterization tasks - symptoms, exposures, 
transmission methods and transmission agents. As can be seen,
{\fullmodel} performs the best for 3 tasks and in average. It is also interesting
to note that {\fullmodel} achieves higher performance gain over
the baseline models for the symptoms category than the other categories. The superior performance 
of {\fullmodel} in the symptoms category can be attributed to two factors - (a) higher ercentage 
of symptom words in the disease vocabulary $\mathcal{V}$ (see Figure~\ref{fig:vocab_dist}) and (b) higher occurrences of symptom words in the
HealthMap news corpus. News articles reporting a disease outbreak generally tend to focus more on the symptoms related to the disease 
rather than the other categories. Given the functionality of Dis2Vec, higher occurrences of symptom terms in outbreak news reports will lead 
to generation of efficient word embeddings for characterizing disease symptoms.}

\begin{table*}[!ht]
  \centering 
  \caption{Comparative performance evaluation of {\fullmodel} with full vocabulary against each of the
  6 conditions of {\fullmodel} with a truncated vocabulary across the 4 characterization tasks where the 
  truncated vocabulary consists of disease names and all possible terms related to a particular taxonomical category. 
  We use equation~\ref{eq:accmetric} as the accuracy metric in this table.}
  \label{tab:eval_vocab}
  
%\scriptsize
\begin{tabular}{|l|l|l|l|l|l|}
  \hline
Characterization tasks       & \begin{tabular}[c]{@{}l@{}}{\fullmodel}\\ (Exposures)\end{tabular} &  \begin{tabular}[c]{@{}l@{}}{\fullmodel}\\ (Transmission methods)\end{tabular}   & \begin{tabular}[c]{@{}l@{}}{\fullmodel}\\ (Transmission agents)\end{tabular}   & \begin{tabular}[c]{@{}l@{}}{\fullmodel}\\ (Symptoms)\end{tabular}   & \begin{tabular}[c]{@{}l@{}}{\fullmodel}\\ (full vocabulary)\end{tabular} \\ 
  \hline
  \hline
  Symptoms            & 0.597 & 0.581 & 0.165 & 0.883  & {\bf 0.945} \\ \hline
  Exposures           & 0.554 & 0.557 & 0.315 & 0.416  & {\bf 0.597} \\ \hline
  Transmission methods & 0.748 & 0.768 & 0.517 & 0.455  & {\bf 0.794} \\ \hline
  Transmission agents & 0.446 & 0.459 & 0.467 & 0.457  & {\bf 0.516}    \\ \hline
\end{tabular}
 
 \end{table*}

\begin{table*}[!ht]
  \centering 
  \caption{Comparison of different parameter settings for each model, measured by the number of characterization tasks 
  in which the best configuration had that parameter setting. Non-applicable combinations are marked by `NA'}
  \label{tab:beneficial_param}
  
%\scriptsize
\begin{tabular}{|l|c|c|c|c|c|cc|}
\hline
\multirow{2}{*}{Method} & \textsf{$T$} & \textsf{$L$} & \textsf{$k$} & \textsf{$\alpha$} & \textsf{$\pi_{s}$} & \textsf{$\pi_{o}$} &  \\
                        & $300:600$ & $5: 10: 15$ & $1:5:15$ & $0.75:1$ & $0.3:0.5:0.7$ & $0.3:0.5:0.7$ &  \\    
\hline
\hline
{\fullmodelall}         &  $2:2$    & $3:1:0$ & $2:1:1$ & $1:3$ & $4:0:0$ & $0:2:2$ &  \\
\hline
{\fullmodelsample}      & $2:2$  & $2:1:1$ & $1:1:2$ & $4:0$ & $1:2:1$ & NA &  \\
\hline
{\fullmodelobjective}   & $3:1$  & $2:2:0$ & $1:1:2$ & $3:1$ & NA & $2:0:2$ &  \\
\hline
{\skipnegative}         & $2:2$  & $2:2:0$ & $0:2:2$ & $2:2$ & NA & NA &  \\
\hline
{\skiphierarchy}        & $3:1$  & $1:0:3$ & NA & NA & NA  & NA &  \\
\hline
{\bagofwords}           & $0:4$  & $0:4:0$ & NA & NA & NA & NA &  \\ 
\hline
\end{tabular}
 \end{table*}

\begin{table}[!ht]
    \centering 
  \caption{Comparative performance evaluation of {\fullmodel} against
  {\skipnegative}, {\skiphierarchy} and {\bagofwords} across the 4 characterization
  tasks for each class of diseases (emerging, endemic and rare) under the best parameter configuration for a particular $\lbrace$disease
  class, task, model$\rbrace$ combination. We use equation~\ref{eq:accmetric} as the accuracy metric in this table.}
  \label{tab:dis_class}
%\scriptsize
\begin{tabular}{|p{2cm}|p{4.7cm}|p{1.8cm}|p{1.7cm}|p{1.7cm}|p{1.90cm}|}
\hline
Class                     & Characterization Tasks & {\bagofwords}  & {\skiphierarchy} & {\skipnegative} & {\fullmodel}        \\ \hline \hline
Emerging                  & Symptoms            & 0.589 & 0.671          & 0.722          & \textbf{0.977} \\ \cline{2-6} 
                          & Exposures           & 0.356 & 0.495          & 0.516          & \textbf{0.679} \\ \cline{2-6} 
                          & Transmission methods & 0.407 & 0.885          & 0.898          & \textbf{0.945}          \\ \cline{2-6} 
                          & Transmission agents & 0.528 & 0.587          & 0.795          & \textbf{0.975} \\ \hline
Endemic                   & Symptoms            & 0.453 & 0.583          & 0.671          & \textbf{0.930} \\ \cline{2-6} 
                          & Exposures           & 0.421 & 0.512          & \textbf{0.642} & 0.631          \\ \cline{2-6} 
                          & Transmission methods & 0.472 & 0.820          & 0.851          & \textbf{0.856} \\ \cline{2-6} 
                          & Transmission agents & 0.164 & 0.399          & 0.408          & \textbf{0.415} \\ \hline
Rare                      & Symptoms            & 0.506 & 0.536          & 0.599          & \textbf{0.949} \\ \cline{2-6} 
                          & Exposures           & 0.377 & 0.525          & 0.616          & \textbf{0.670} \\ \cline{2-6} 
                          & Transmission methods & 0.503 & 0.760          & 0.755          & \textbf{0.775} \\ \cline{2-6} 
                          & Transmission agents & 0.320 & \textbf{0.522} & 0.512          & 0.515          \\ \hline
\end{tabular}
 \end{table}

\par{\textbf{What are beneficial parameter configurations for characterizing 
    diseases?} To identify which parameter settings are beneficial for characterizing diseases, 
    we looked at the best parameter configuration of all the 6 models on each task. We then counted the number of times each parameter setting was chosen in 
    these configurations (see Table~\ref{tab:beneficial_param}). We compared standard settings of each parameter as explored in previous research~\cite{levyparam}. For 
    the new parameters $\pi_{s}$ and $\pi_{o}$ introduced by {\fullmodel}, we chose the values $0.3$, $0.5$ and $0.7$ in order to analyze the impact of these parameters 
    with values $< 0.5$ and $\geq 0.5$. For {\fullmodelobjective} and {\fullmodelall}, some trends emerge regarding 
    the parameter $\pi_{o}$ that these two models consistently benefit from values of $\pi_{o} \geq 0.5$
    validating our claims in section~\ref{sec:our_model} that when $\pi_{o} \geq 0.5$, disease words and non-disease
    words get scattered from each other in the vector space, thus tending to generate disease specific embeddings.
    However, for $\pi_{s}$ we observe mixed trends. As expected, {\fullmodelsample} benefits from higher values of sampling 
    parameter $\pi_{s} \geq 0.5$. But {\fullmodelall} seems to prefer lower values of $\pi_{s} \textless 0.5$ and higher values of
    $\pi_{o} \geq 0.5$ for the disease characterization tasks. For the smoothing parameter($\alpha$), all the applicable models prefer 
    smoothed unigram distribution ($\alpha=0.75$) for negative sampling except {\fullmodelall} which is in favor of unsmoothed
    distribution ($\alpha=1.0$) for characterizing diseases. For the number of \textit{negative} samples $k$, all the applicable
    models seem to benefit from $k \textgreater 1$ except {\fullmodelall} which seems to prefer $k = 1$. For the window size ($L$), all the models prefer smaller-sized
    context windows (either 5 or 10) except {\skiphierarchy} which prefers larger-sized windows ($L \textgreater 10$) for characterizing diseases. Finally, regarding the
    dimensionality ($T$) of the embeddings, {\fullmodelall}, {\fullmodelsample} and {\skipnegative} are in equal favor of both 300 and 600 dimensions. {\fullmodelobjective} 
    and {\skiphierarchy} prefer 300 dimensions and {\bagofwords} is in favor of 600 dimensions for characterizing diseases.}

\par{\textbf{Importance of taxonomical categories - how should we 
construct the disease vocabulary?}
We followup our previous analysis by investigating the importance of words related to 
each taxonomical category in constructing the disease vocabulary towards final characterization accuracy.
To evaluate a particular category, we used a truncated disease vocabulary consisting of disease names and 
the words in the corresponding category to drive the discovery of word embeddings in {\fullmodel} under the 
best parameter configuration for that category.
We compared the accuracy of each of these conditions ({\fullmodel} (exposures), {\fullmodel} (transmission methods), 
{\fullmodel} (transmission agents),
{\fullmodel} (symptoms)) against {\fullmodel} (full vocabulary) across the 4 characterization tasks. Table~\ref{tab:eval_vocab} presents our results for this analysis 
and provides multiple insights as follows. (a) Constructing the vocabulary with words related to all the categories leads to better characterization across all the tasks.
(b) As expected, $\fullmodel$ (symptoms) is the second best performing model for the symptoms category 
but it's performance is degraded for other tasks. The same goes for {\fullmodel} (transmission methods) and {\fullmodel} (transmission agents).
(c) Therefore, it indicates that in order to achieve reasonable characterization accuracy for a category, we need to supply at least the words related to that category along with the disease names in constructing the vocabulary.}

\begin{figure*}[t!]
  \centering
  {\large{\bf H7N9} \hspace{10.5em} {\bf Avian influenza} \hspace{10.5em}{\bf Plague}}
  \vspace{3.0em}
  \begin{subfigure}{0.33\textwidth}
    \centering
    \includegraphics[width=1.0\linewidth]{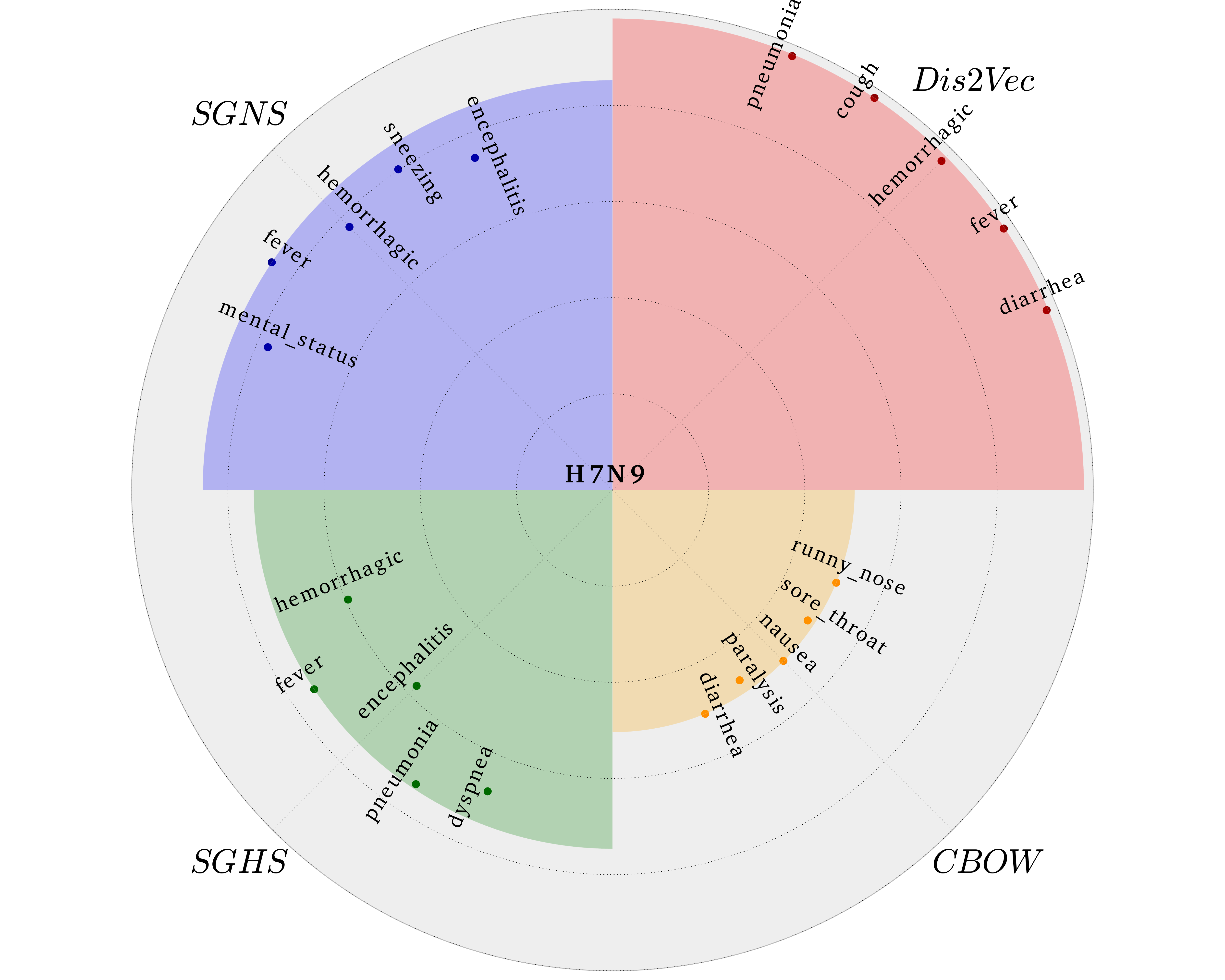}
    \caption{Symptoms}
    \label{fig:ebola:symptoms}
  \end{subfigure}
  \hspace{-1em}
  \begin{subfigure}{0.33\textwidth}
    \centering
    \includegraphics[width=1.0\linewidth]{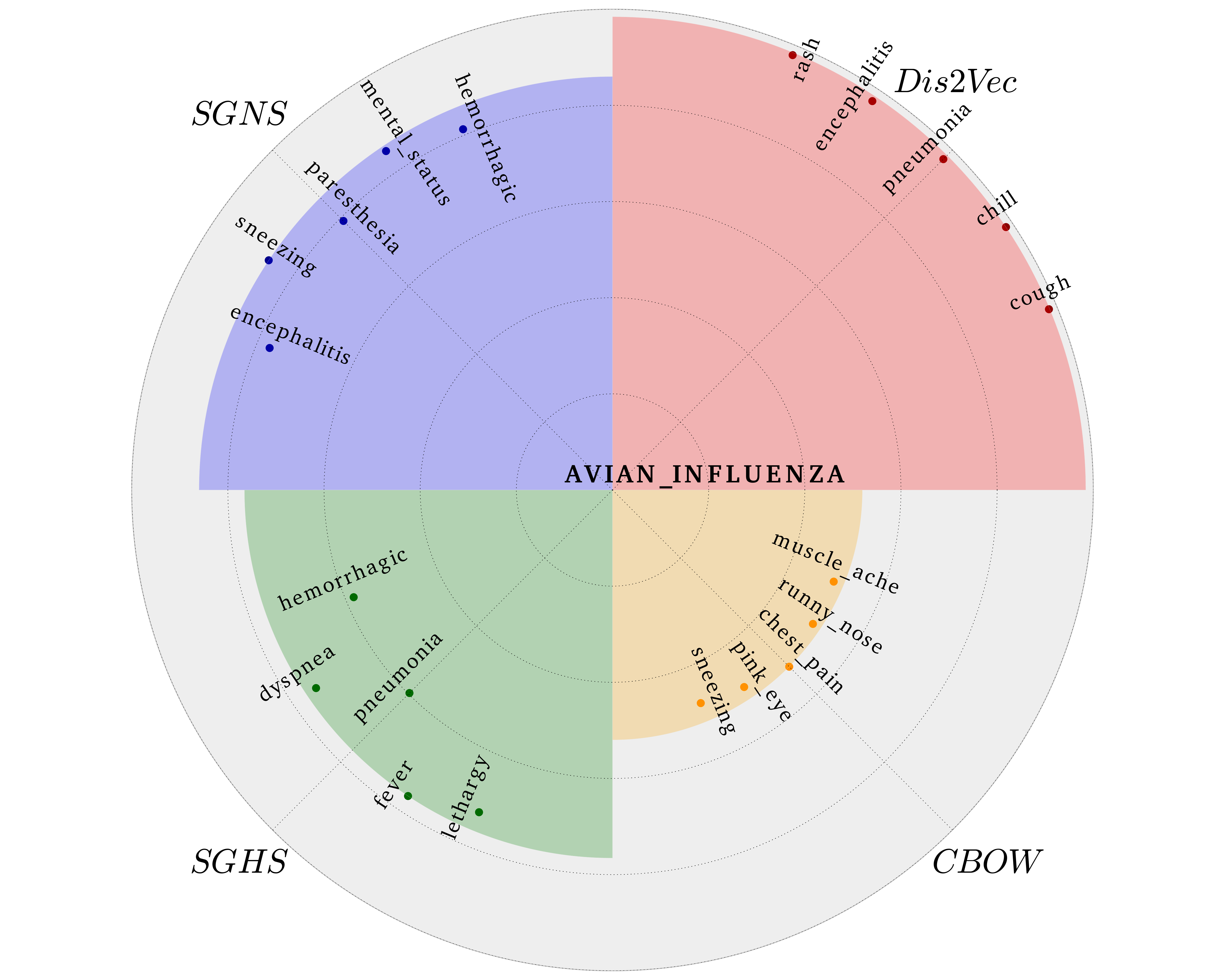}
    \caption{Symptoms}
    \label{fig:mers:symptoms}
  \end{subfigure}
  \hspace{-1em}
  \begin{subfigure}{0.33\textwidth}
    \centering
    \includegraphics[width=1.0\linewidth]{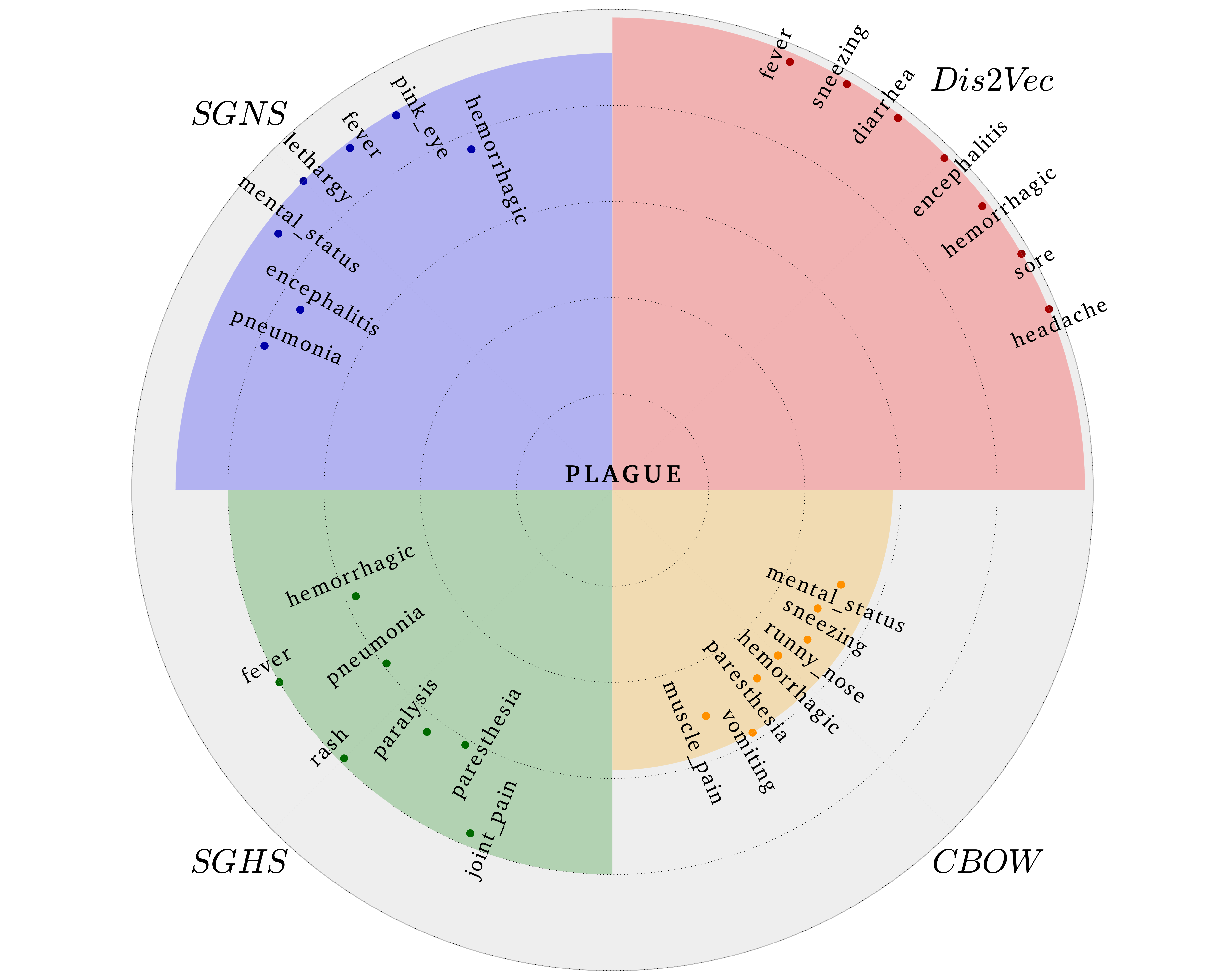}
    \caption{Symptoms}
    \label{fig:h7n9:symptoms}
  \end{subfigure}
  \\
  \vspace{-2em}
  \begin{subfigure}{0.33\textwidth}
    \centering
    \includegraphics[width=1.0\linewidth]{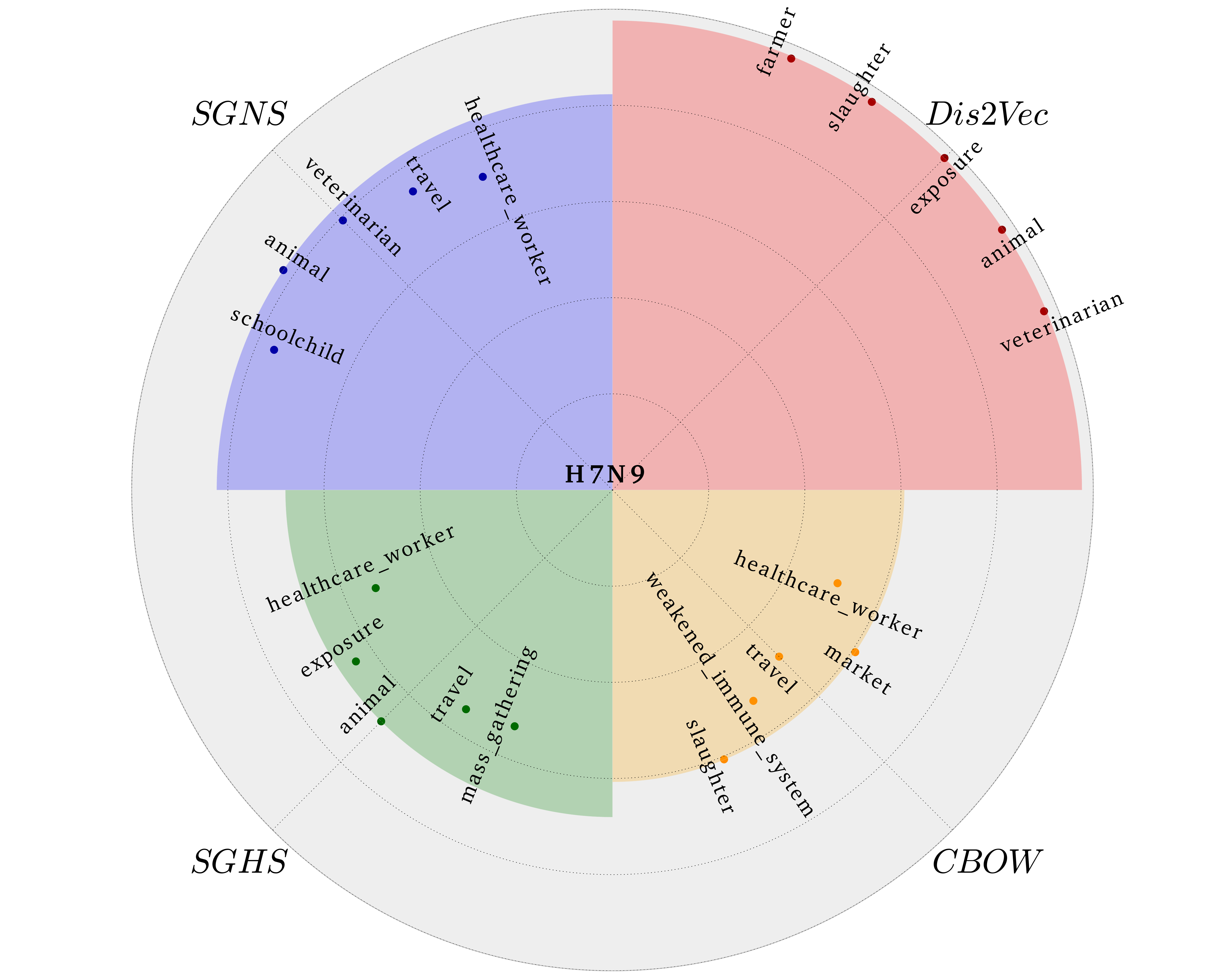}
    \caption{Exposures}
    \label{fig:ebola:exposures}
  \end{subfigure}
  \hspace{-1em}
  \begin{subfigure}{0.33\textwidth}
    \centering
    \includegraphics[width=1.0\linewidth]{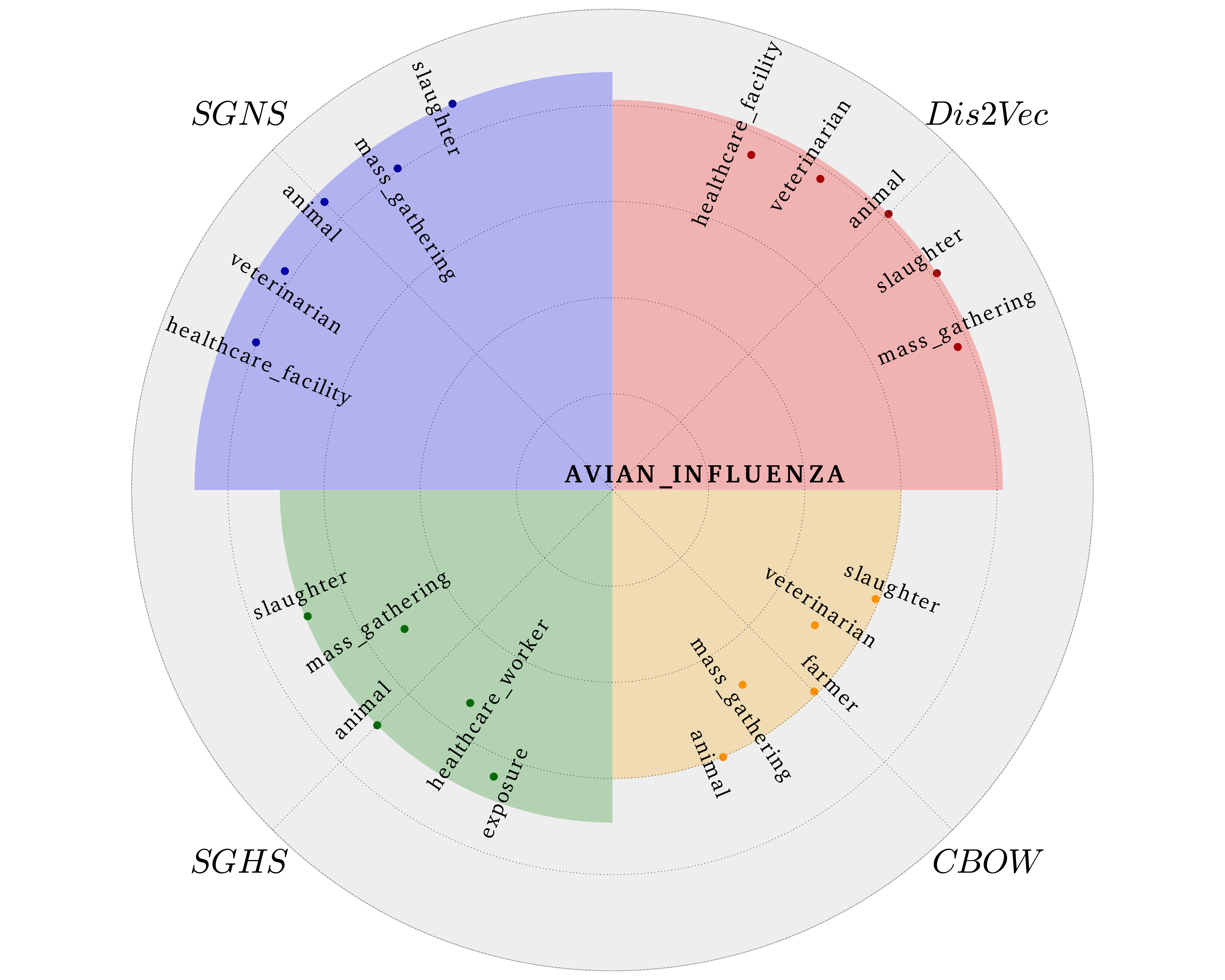}
    \caption{Exposures}
    \label{fig:mers:exposures}
  \end{subfigure}
  \hspace{-1em}
  \begin{subfigure}{0.33\textwidth}
    \centering
    \includegraphics[width=1.0\linewidth]{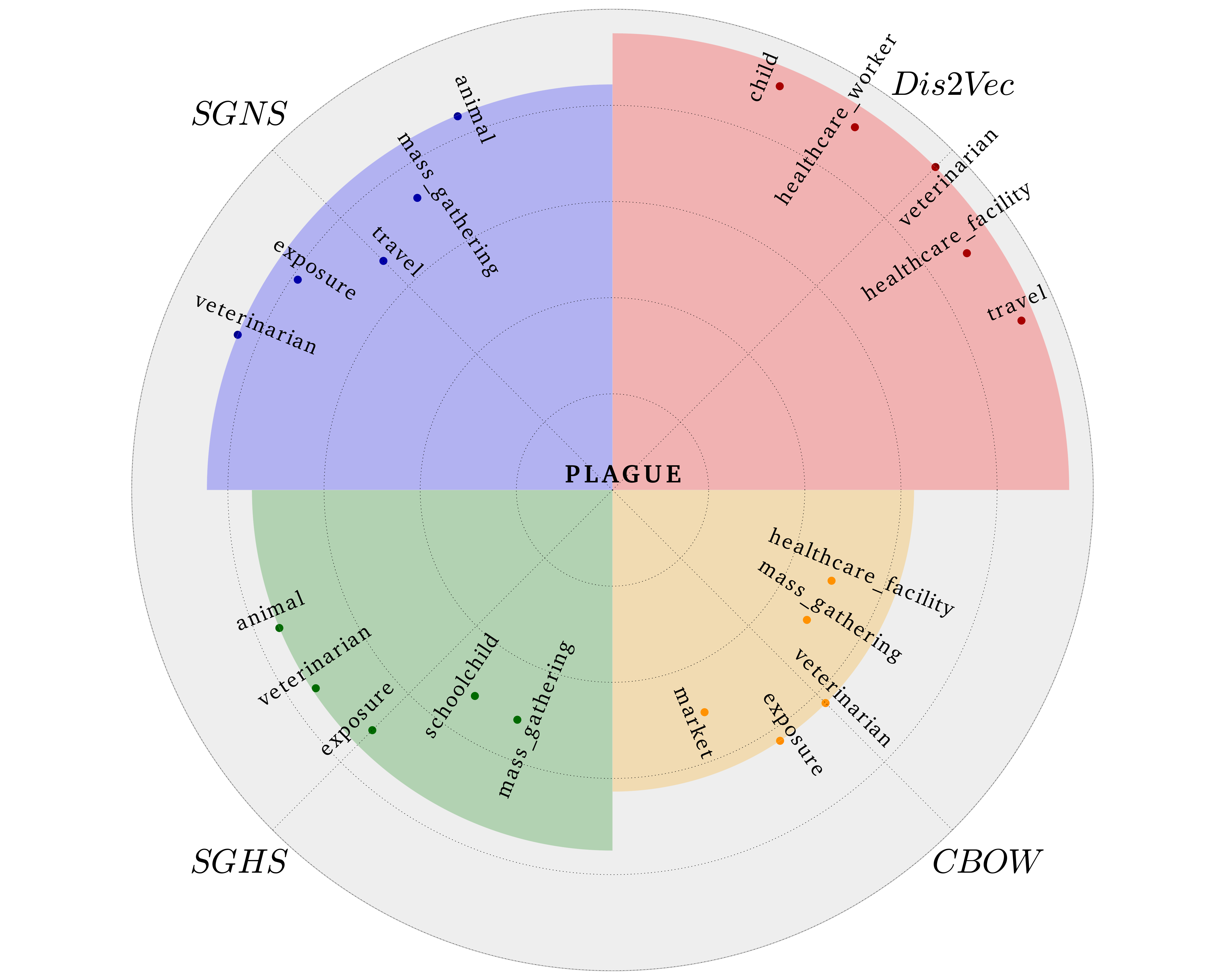}
    \caption{Exposures}
    \label{fig:h7n9:exposures}
  \end{subfigure}
  \\
    \begin{subfigure}{0.33\textwidth}
    \centering
    \includegraphics[width=1.0\linewidth]{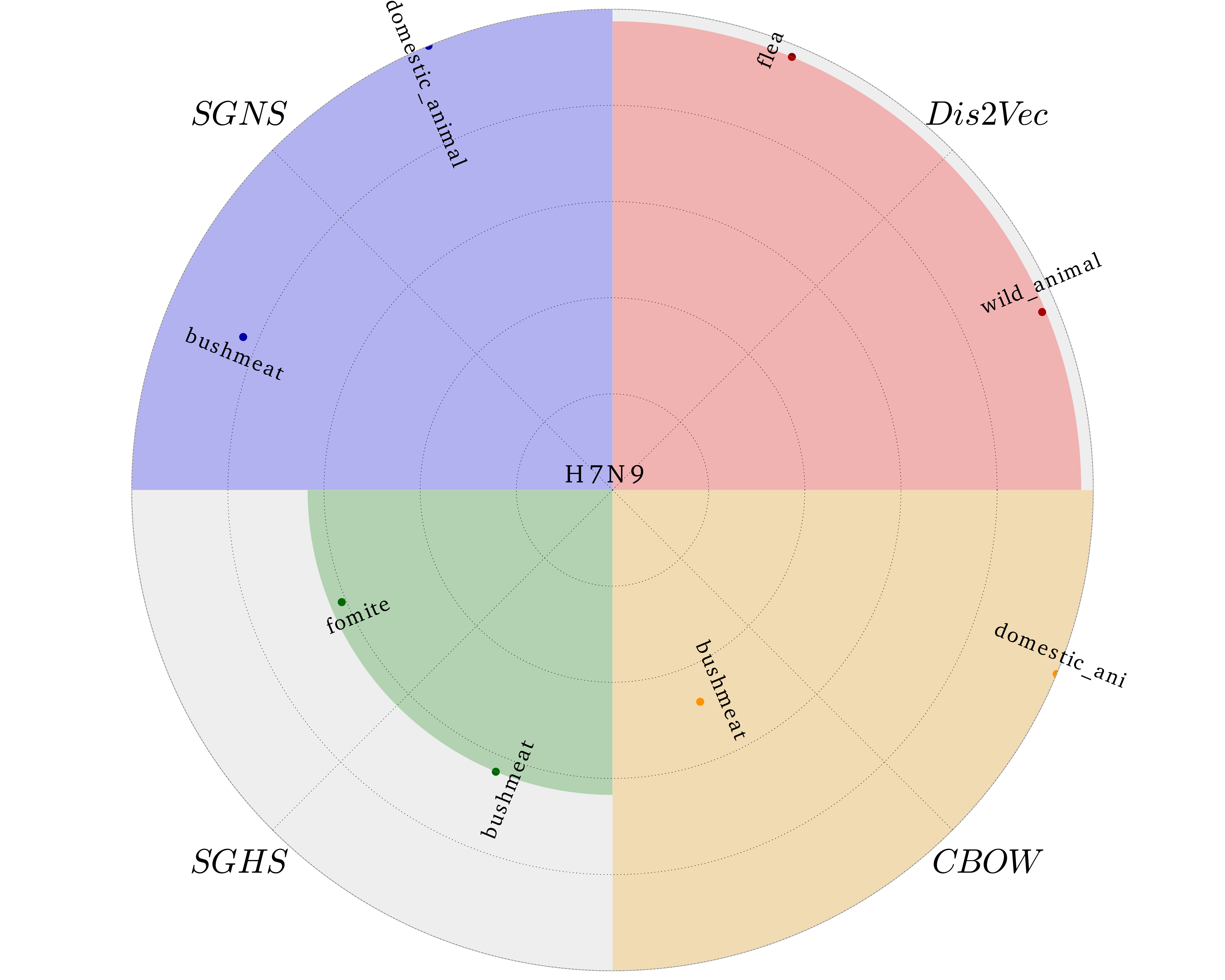}
    \caption{Transmission Agents}
    \label{fig:ebola:tx_agents}
  \end{subfigure}
  \hspace{-1em}
  \begin{subfigure}{0.33\textwidth}
    \centering
    \includegraphics[width=1.0\linewidth]{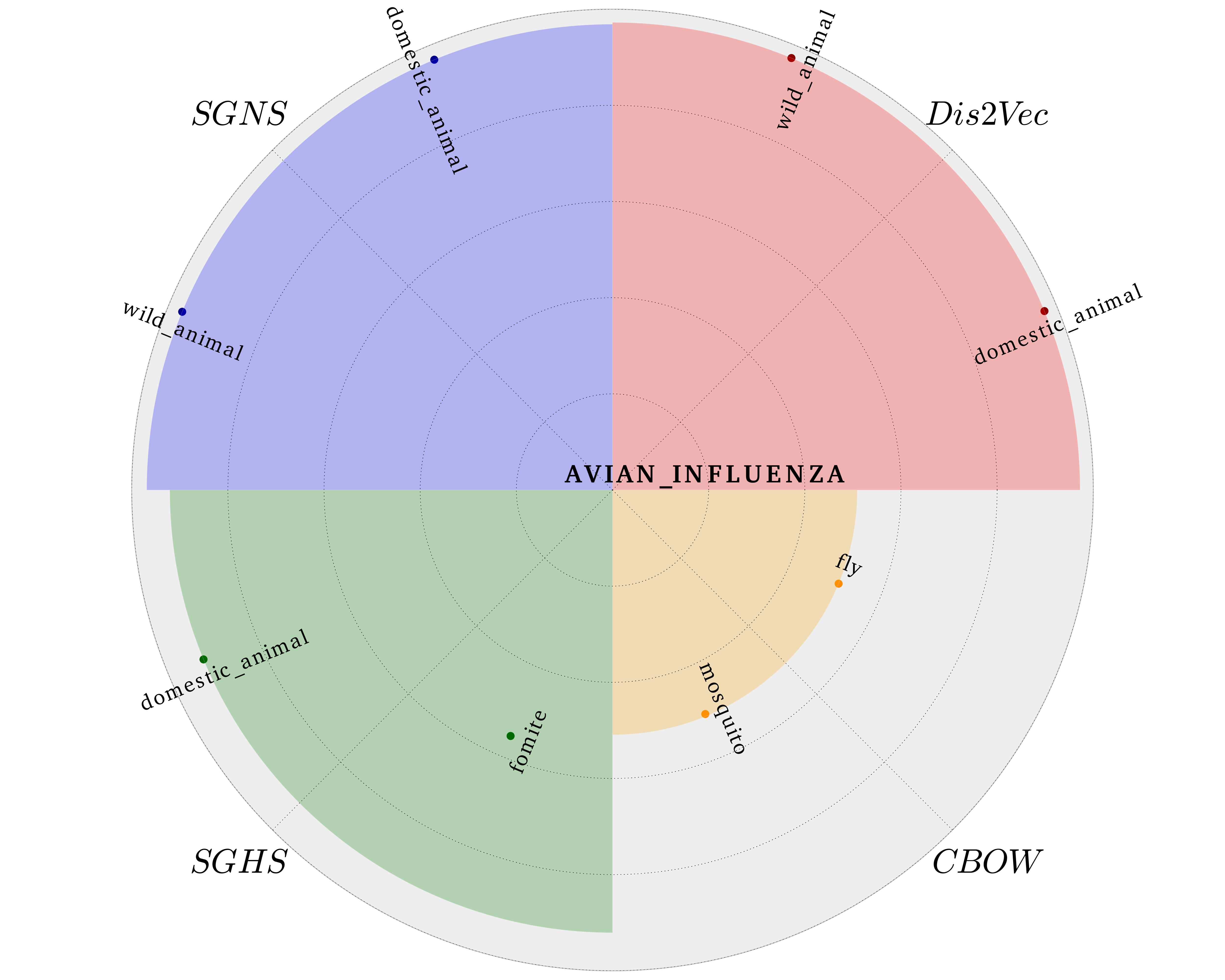}
    \caption{Transmission Agents}
    \label{fig:mers:tx_agents}
  \end{subfigure}
  \hspace{-1em}
  \begin{subfigure}{0.33\textwidth}
    \centering
    \includegraphics[width=1.0\linewidth]{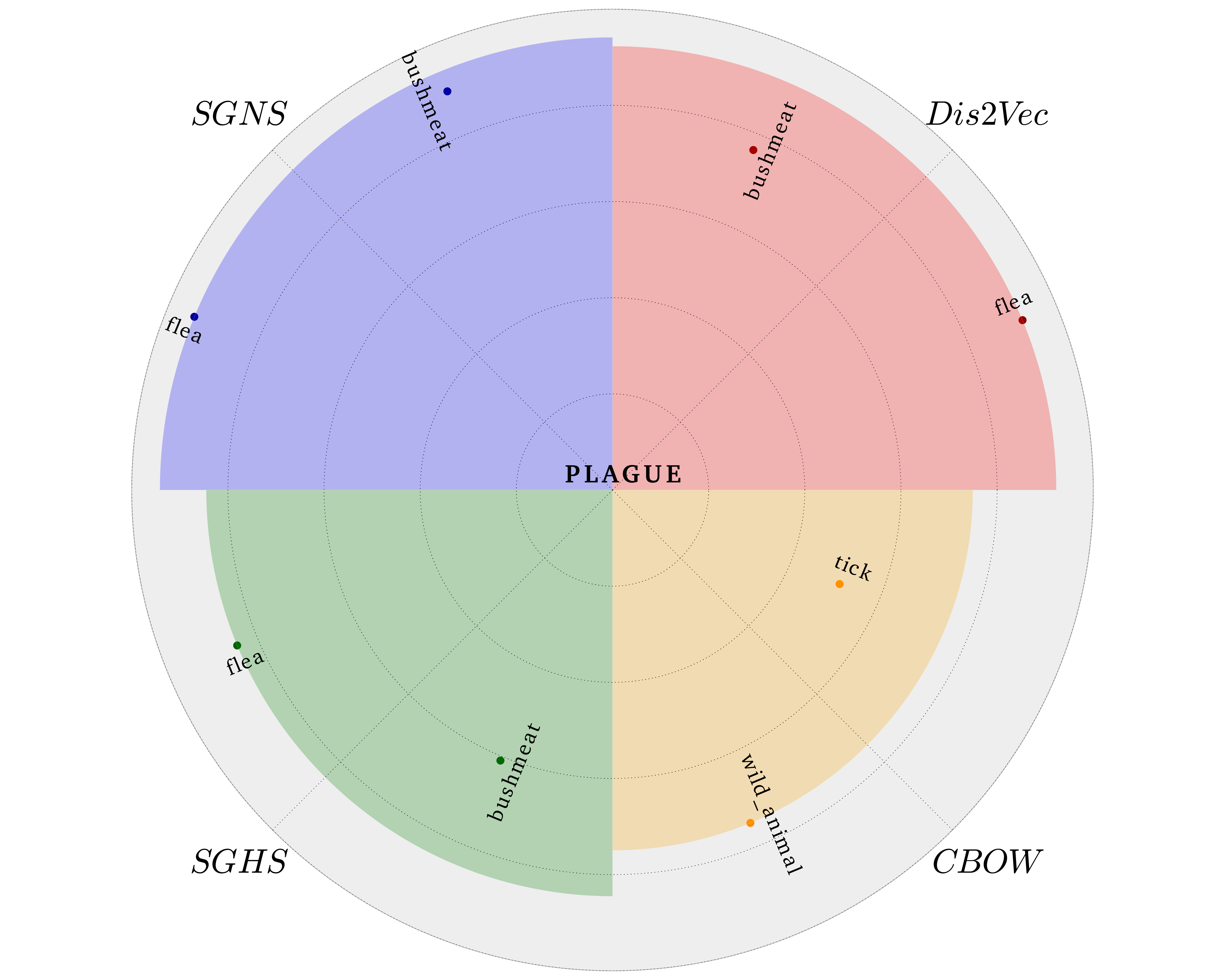}
    \caption{Transmission Agents}
    \label{fig:h7n9:tx_agents}
  \end{subfigure}
  \\
    \begin{subfigure}{0.33\textwidth}
    \centering
    \includegraphics[width=1.0\linewidth]{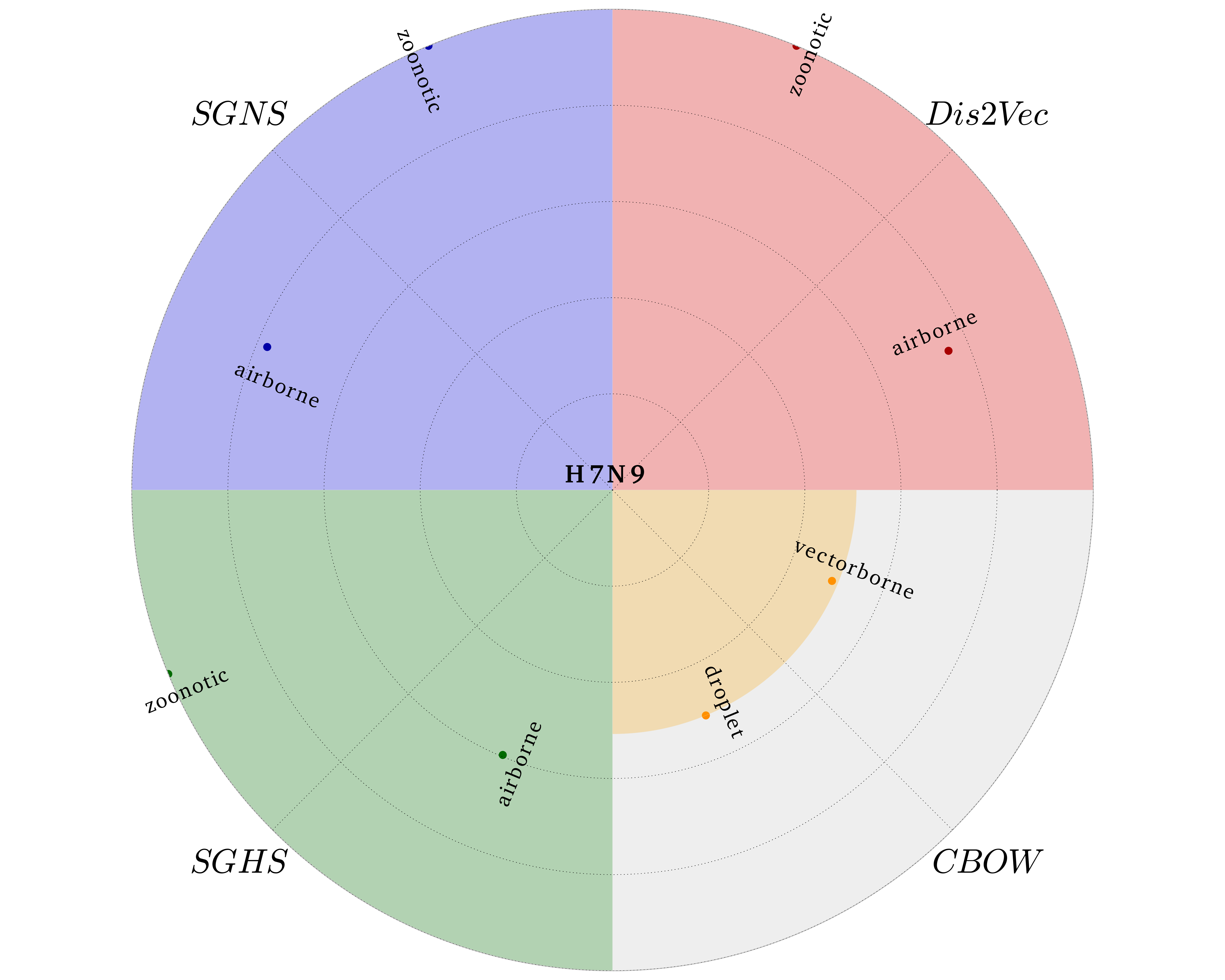}
    \caption{Transmission Method}
    \label{fig:ebola:tx_nature}
  \end{subfigure}
  \hspace{-1em}
  \begin{subfigure}{0.33\textwidth}
    \centering
    \includegraphics[width=1.0\linewidth]{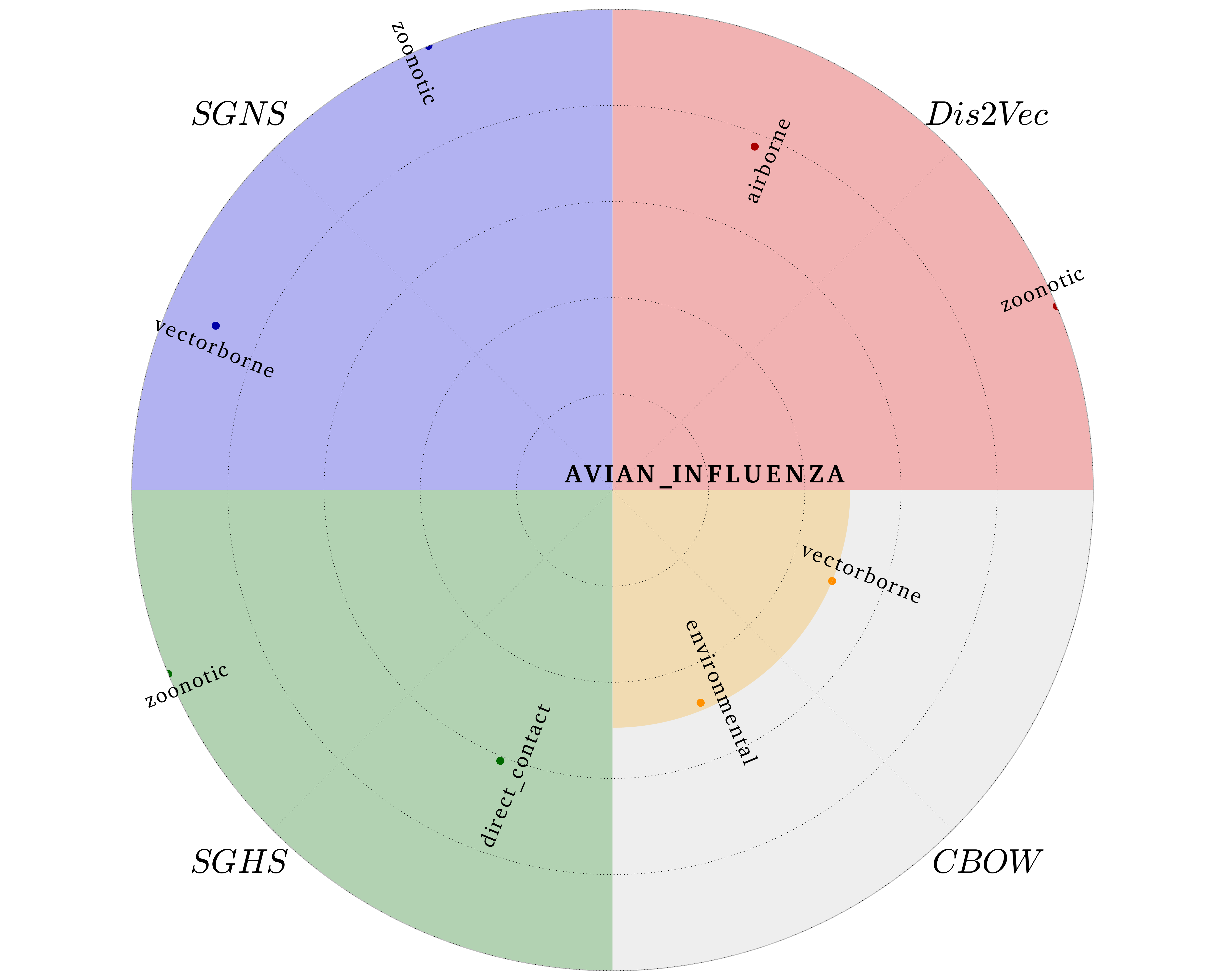}
    \caption{Transmission Method}
    \label{fig:mers:tx_nature}
  \end{subfigure}
  \hspace{-1em}
  \begin{subfigure}{0.33\textwidth}
    \centering
    \includegraphics[width=1.0\linewidth]{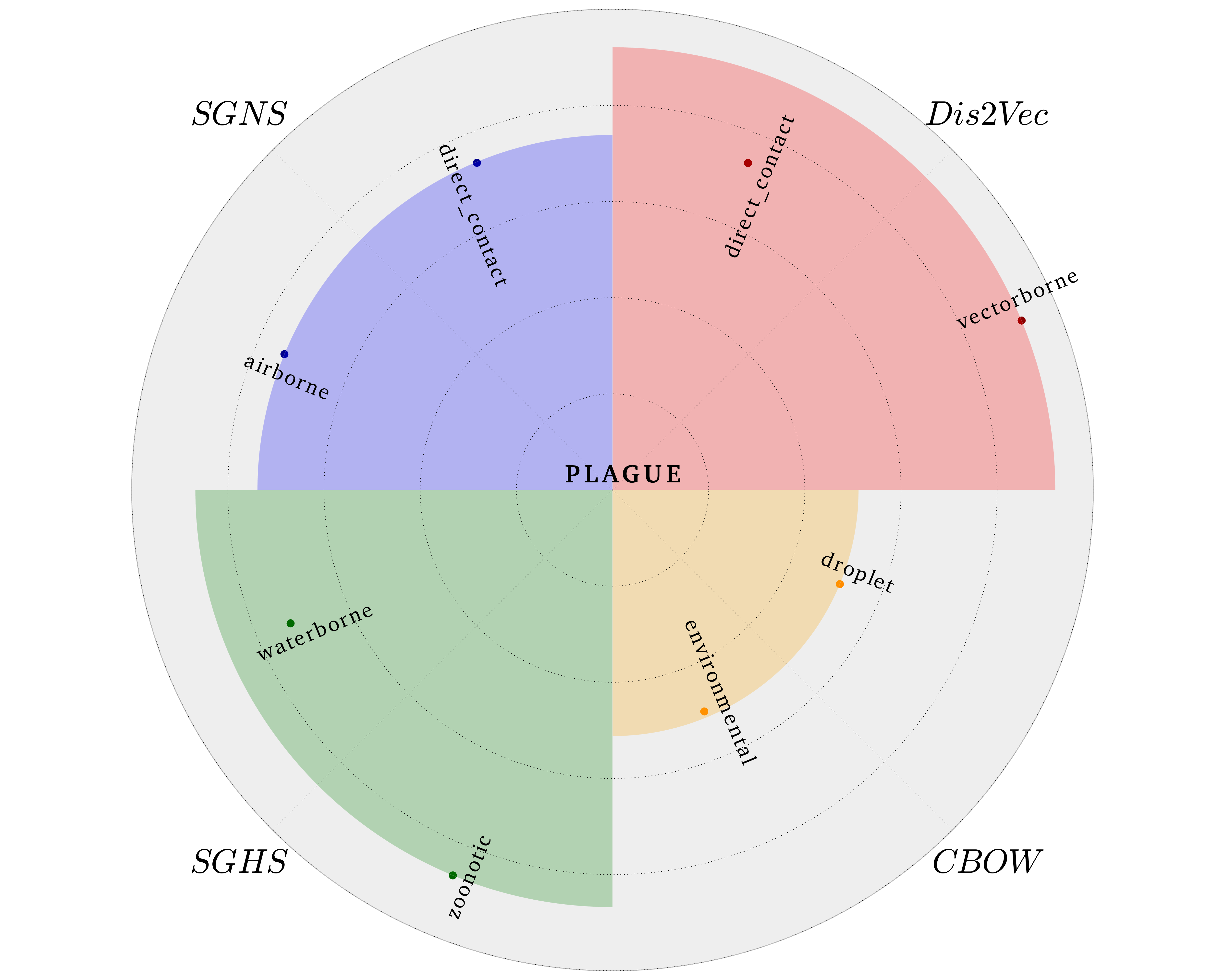}
    \caption{Transmission Method}
    \label{fig:h7n9:tx_nature}
  \end{subfigure}
  \\
  \caption{Case study for emerging, endemic and rare diseases: Disease characterization accuracy
  plot for {\fullmodel}~(first quadrant, red), \skipnegative~(second quadrant, blue), 
  \skiphierarchy~(third quadrant, green),  and \bagofwords~(fourth quadrant, orange)
  w.r.t.\ H7N9 (left, emerging), avian influenza (middle, endemic) and plague (right, rare).
  The shaded area in a quadrant indicates the cosine similarity (scaled between
  0 and 1) of the top words found for the category of interest using 
  corresponding model, as evaluated against the human annotated words (see Table~\ref{tab:taxonomy_examples}).
  The top words found for each model is shown in the corresponding
  quadrant with radius equal to its average similarity 
  with the human annotated  words for the disease.
  {\fullmodel} shows best overall performance with noticeable improvements for
  symptoms w.r.t.\ all diseases.}
  \label{fig:case_study}
\end{figure*}

\par{\textbf{Can {\fullmodel} be applied to characterize emerging, endemic and rare diseases? }
We classified the 39 diseases of interest into 3 classes as follows. For classifying each disease,
we plotted the time series of the counts of HealthMap articles with disease tag equal to the corresponding 
disease.

\noindent $\bullet$ \textbf{Endemic}: We considered a disease as endemic if the counts of articles were consistently high
    for all years with repeating shapes. E.g.- rabies, avian influenza, west nile virus.\\
\noindent $\bullet$  \textbf{Emerging}: We considered a disease as emerging if the counts of articles were 
                historically low, but have peaked in recent years. E.g.- Ebola, H7N9, MERS.\\
\noindent $\bullet$  \textbf{Rare}: We considered a disease as rare if the counts were consistently low for all years with or 
    without sudden spikes. E.g.- plague, chagas, japanese encephalitis. We also considered a disease as rare 
    if the counts of articles were high in 2010/2011, but have since fallen down and depicted consistently low counts. E.g.- tuberculosis. \\

  Following classification, we found 4 emerging diseases, 12 endemic diseases and 
  23 rare diseases. The full list of the diseases under each class and the time series plots of the HealthMap article counts for these diseases 
  can be accessed at $\href{https://github.com/sauravcsvt/Dis2Vec_supplementary}{Dis2Vec-supplementary}$. In Table~\ref{tab:dis_class}, we compared the
  accuracy of {\fullmodel} against the baseline word2vec models for each class of diseases across the 4 characterization tasks under
  the best parameter configuration for a particular $\lbrace$disease class, task, model$\rbrace$ combination. It can be seen that {\fullmodel} is
  the best performing model for majority of the $\lbrace$disease class, task$\rbrace$ combinations except $\lbrace$endemic, exposures$\rbrace$
  and $\lbrace$rare, transmission agents$\rbrace$. It is interesting to note that for the symptoms category, {\fullmodel} performs better than 
  the baseline models across all the disease classes. Irrespective of disease class, news reports generally 
  mention the symptoms of the disease while reporting an outbreak. As the characteristics of the emerging diseases are relatively unknown w.r.t.\
  endemic and rare, news media reports also tend to focus on other categories (exposures,
  transmission methods, transmission agents) apart from the symptoms to create awareness among the general public. Therefore, {\fullmodel} performs 
  better than the baseline models across all the categories for the emerging diseases. For endemic and rare diseases, {\fullmodel} outperforms the 
  baseline models w.r.t.\ the symptoms category. For other categories, {\fullmodel} performs better overall, although the performance gain is not high
  in comparison to the symptoms. News media reports tend to neglect categories other than symptoms while reporting
  outbreaks of endemic and rare diseases. In Figure~\ref{fig:case_study}, we show the top words found for each category of an emerging disease (H7N9), an 
  endemic disease (avian influenza) and a rare disease (plague) across all the models. The human annotated words corresponding to each category of these diseases 
  can be found in Table~\ref{tab:taxonomy_examples}. We selected these 3 diseases due to their public health significance and the fact that these diseases have complete 
  coverage across all the taxonomical categories (see Table~\ref{tab:taxonomy_examples}). It is interesting to note that for H7N9, the top words found by {\fullmodel} 
  for the symptoms category contain all
  the human annotated words \textit{fever}, \textit{cough} and \textit{pneumonia}, while the top words found by {\skipnegative} only contain the word \textit{fever}.
  For exposures (H7N9), {\fullmodel} is able to capture three human annotated words \textit{animal exposure}, {farmer}, 
  {slaughter}. However, {\skipnegative} is only able to capture the word \textit{animal}. For the symptoms category of the rare disease plague, {\fullmodel} is 
  able to detect three human annotated words \textit{sore}, \textit{fever} and \textit{headache} with {\skipnegative} only being able to detect the word \textit{fever}.
  Moreover, {\fullmodel} is able to characterize the transmission method of plague as \textit{vectorborne} with {\skipnegative} failing to do so.}

\section{Conclusions}
  \label{sec:concl}
  
Classical word2vec methods such as {\skipnegative} and {\skiphierarchy}
have been applied to solve a variety of linguistic tasks with considerable
accuracy.  
However, such methods fail to generate satisfactory embeddings for highly
specific domains such as healthcare where uncovering the relationships with respect
to domain specific words is of greater importance than the non-domain ones.
These algorithms are by design unsupervised and do not
permit the inclusion of domain information to find interesting embeddings.
In this paper, we have proposed {\fullmodel}, 
a disease specific word2vec framework that given an unstructured news 
corpus and domain knowledge in terms of important words, can 
find interesting disease characterizations. We demonstrated the strength of our
model by comparing it against three classical word2vec methods on
four disease characterization tasks. {\fullmodel} exhibits the best overall
accuracy for 3 tasks across all the diseases and in general, its relative performance improvement
is found to be empirically dependent on the amount of supplied domain 
knowledge. Consequently, {\fullmodel} works
especially well for characteristics with more domain knowledge (symptoms)
and is found to be a promising tool to analyze different class of diseases viz. emerging, endemic 
and rare. In future, we aim to analyze a greater variety of diseases and try 
to ascertain common relationships between such diseases across different
geographical regions.

\section{Supplementary Data}
  The disease specific word embeddings generated by {\fullmodel} from the HealthMap
  corpus can be accessed at \href{https://github.com/sauravcsvt/Dis2Vec_supplementary}{Dis2Vec-supplementary}.

\iffalse
\section{Acknowledgements}
{ Supported by the Intelligence Advanced Research Projects Activity
  (IARPA) via Department of Interior National Business Center (DoI/NBC)
  contract number D12PC000337, the US Government is authorized to
  reproduce and distribute reprints of this work for Governmental
  purposes notwithstanding any copyright annotation thereon.
  Disclaimer: The views and conclusions contained herein are those of
  the authors and should not be interpreted as necessarily representing
  the official policies or endorsements, either expressed or implied, of
  IARPA, DoI/NBC, or the US Government.
}
\fi

\bibliographystyle{abbrv}

\end{document}